\pdfoutput=1

\documentclass[10pt]{article}

\def\month{08}
\def\year{2026}
\def\openreview{\url{https://openreview.net/forum?id=I9UjKxW4nq}}
\usepackage[accepted]{tmlr}

\usepackage{amsmath,amsfonts,bm}

\def\eqref#1{equation~\ref{#1}}

\def\1{\bm{1}}

\DeclareMathAlphabet{\mathsfit}{\encodingdefault}{\sfdefault}{m}{sl}
\SetMathAlphabet{\mathsfit}{bold}{\encodingdefault}{\sfdefault}{bx}{n}

\usepackage{fontawesome5}

\usepackage{latexsym}
\usepackage{multirow}
\usepackage[normalem]{ulem}
\usepackage{url}
\usepackage[T1]{fontenc}
\usepackage{pythonhighlight}
\usepackage{amsmath}
\usepackage{verbatim}
\usepackage{paralist}
\usepackage{framed}
\usepackage{xcolor}
\usepackage{graphicx}
\usepackage{wrapfig}
\usepackage{subcaption}
\usepackage[most]{tcolorbox}
\usepackage[utf8]{inputenc}
\usepackage{enumitem}
\usepackage{CJKutf8}
\usepackage[normalem]{ulem}
\usepackage{hyperref}
\hypersetup{
    colorlinks=true,
    linkcolor=blue,
    citecolor=blue,
    urlcolor=blue,
    pdftitle={AlignSAE: Concept-Aligned Sparse Autoencoders}
}
\usepackage[capitalize]{cleveref}
\usepackage{microtype}

\usepackage{inconsolata}
\usepackage{soul}
\usepackage{amssymb}
\usepackage{booktabs}
\usepackage{graphicx}
\usepackage[export]{adjustbox}
\usepackage{soul}
\usepackage{xcolor}
\usepackage{soulpos} 
\usepackage{etoolbox}
\usepackage{colortbl}
\usepackage{siunitx}
\usepackage{tabularx}
\usepackage{icomma}
\usepackage{pifont}

\usepackage{pgfplots}
\pgfplotsset{compat=1.18}
\usepgfplotslibrary{groupplots}
\usepackage{subcaption}
\usepackage{siunitx}
\definecolor{metricS}{HTML}{1f77b4} 
\definecolor{metricP}{HTML}{ff7f0e} 
\definecolor{metricE}{HTML}{8c564b} 
\definecolor{findinggreen}{HTML}{E8F9E8}
\usepackage{algorithm}
\usepackage{algorithmicx}
\usepackage{algpseudocode}
\usepackage{placeins}
\usepackage{needspace}
\usepackage{amsfonts}  

\usepackage{graphicx}
\usepackage{booktabs}
\usepackage{makecell}
\definecolor{myorange}{HTML}{dd614e}
\definecolor{myyellow}{HTML}{876e2f}
\usepackage{array, booktabs, ragged2e}
\newcolumntype{R}[1]{>{\RaggedLeft\arraybackslash}p{#1}}
\newcolumntype{L}[1]{>{\RaggedRight\arraybackslash}p{#1}}
\usepackage{pifont}
\newcommand{\cmark}{{\ding{51}}}
\newcommand{\xmark}{{\ding{55}}}

\definecolor{mwpblue}{HTML}{0171C8}
\definecolor{mwpred}{HTML}{EF5544}
\definecolor{mwpgreen}{HTML}{00A676}
\definecolor{posdelta}{RGB}{0,128,0} 

\usepackage{xspace}
\usepackage{amsthm}
\theoremstyle{definition}

\newcommand{\revision}[1]{#1}
\newcommand{\resicon}[2]{\href{#1}{#2}}

\crefformat{section}{\S#2#1#3} 
\crefformat{subsection}{\S#2#1#3}
\crefformat{subsubsection}{\S#2#1#3}
\crefmultiformat{section}{\S#2#1#3}{, #2#1#3}{, #2#1#3}{, #2#1#3}

\usepackage{seqsplit}

\newcommand{\modelname}{\textsc{{AlignSAE}}}
\newtcolorbox{findingbox}[1][]{
    breakable,
    enhanced,
    sharp corners,
    boxrule=0pt,
    colback=findinggreen,
    colframe=findinggreen,
    frame hidden,
    borderline west={2pt}{0pt}{green!70!black},
    left=6pt,
    right=6pt,
    top=4pt,
    bottom=4pt,
    before skip=10pt,
    after skip=10pt,
    fontupper=\linespread{1.0}\selectfont,
    #1
}

\newtcolorbox{resultbox}{
  breakable,
  sharp corners,
  colframe=black,
  colback=white,
  boxrule=0.4pt,
  left=2pt,
  right=2pt,
  top=2pt,
  bottom=2pt,
  fontupper=\linespread{0.7}\selectfont
}

\usepackage{changepage}

\title{
\raisebox{-0.3\height}
{
\includegraphics[width=2em,height=2em]{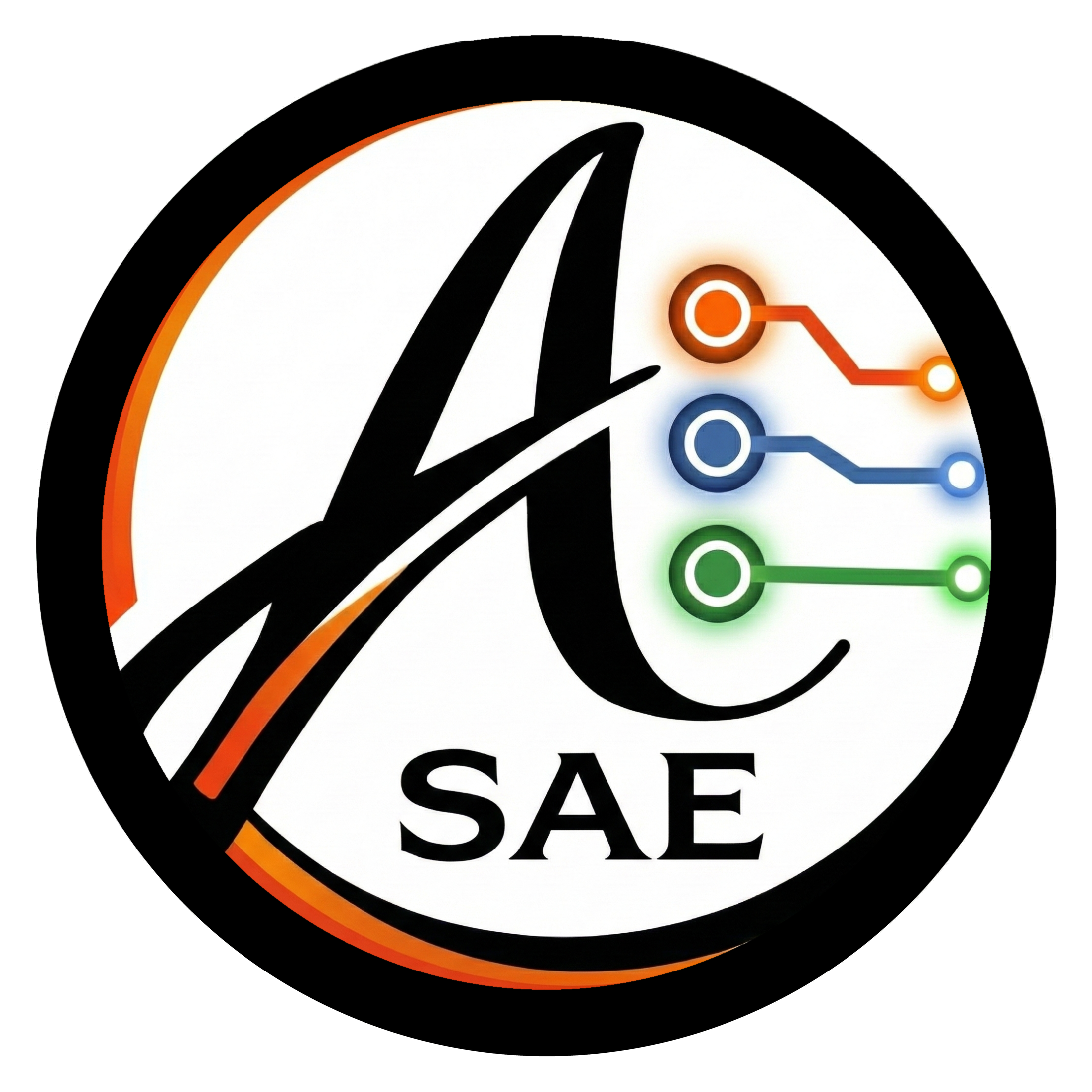}}\xspace
\modelname{}: Concept-Aligned Sparse Autoencoders
}

\author{\name Minglai Yang\thanks{\ Corresponding authors.}\ \thanks{\ Work done while MY was a student at the University of Arizona.} \email mingly@arizona.edu \\
      \addr University of Arizona
      \AND
      \name Xinyu Guo \email xinyuguo@arizona.edu \\
      \addr University of Arizona
      \AND
      \name Zhengliang Shi \email zhenglis@andrew.cmu.edu \\
      \addr Carnegie Mellon University
      \AND
      \name Jinhe Bi \email bijinhe@outlook.com \\
      \addr LMU Munich
      \AND
      \name Steven Bethard \email bethard@arizona.edu \\
      \addr University of Arizona
      \AND
      \name Mihai Surdeanu\footnotemark[2] \email msurdeanu@arizona.edu \\
      \addr University of Arizona
      \AND
      \name Liangming Pan\footnotemark[2] \email liangmingpan@pku.edu.cn \\
      \addr Peking University
}

\begin{document}

\setcounter{footnote}{1}
\maketitle

\begin{center}
\resicon{https://github.com/yminglai/AlignSAE}{\faGithub\ \textbf{Code \& Data}}
\quad
\resicon{https://ymingl.com/assets/pdf/AlignSAEslides.pdf}{\faTv\ \textbf{Slides}}
\quad
\resicon{https://ymingl.com/alignsae-site/}{\faGlobe\ \textbf{Website}}
\end{center}

\begin{abstract}
Large Language Models (LLMs) encode factual knowledge within hidden parametric spaces that are difficult to inspect or control. While Sparse Autoencoders (SAEs) can decompose hidden activations into more fine-grained, interpretable features, they often struggle to reliably align these features with human-defined concepts, resulting in entangled and distributed feature representations.
To address this, we introduce \modelname{},
a method that aligns SAE features with a predefined ontology through a ``pre-train, then post-train'' curriculum. After an initial unsupervised training phase, we apply supervised post-training to bind specific concepts to dedicated latent slots while preserving the remaining capacity for general reconstruction. This separation creates an interpretable interface where specific concepts can be inspected and controlled without interference from unrelated features.
Empirical results on GPT-2 across synthetic biographical QA
(1-hop, 6 concepts) and compositional relational reasoning
(2-hop, 20 concepts) demonstrate that \modelname{} enables
precise causal interventions, such as reliable concept swaps
(94.8\% success in 1-hop), by targeting single, semantically
aligned slots. \modelname{} achieves disentanglement improvements
of $9$--$\infty\times$ over standard SAEs (measured via Mutual
Information Gap), and further supports multi-hop reasoning and
a mechanistic probe of grokking-like generalization dynamics.
Although our evaluation focuses on GPT-2, \modelname{} itself is model-agnostic.
\end{abstract}

\section*{Errata}
This arXiv version clarifies a few issues in the TMLR version:
\begin{itemize}[leftmargin=*,itemsep=1pt,topsep=2pt]
    \item The concept--slot binding metric is now consistently called \emph{diagonal accuracy} (\cref{subsec:metrics}, \cref{app:metrics}); the two earlier definitions of $\mathrm{Acc}_{\text{binding}}$ and the unused top-$k$ accuracy were removed.
    \item The caption and legends of \cref{fig:combined_binding_swap} now state the data split behind each curve (left: training set and unseen-template test split; right: held-out test questions).
\end{itemize}

    \section{Introduction}
\label{sec:introduction}

While Large Language Models (LLMs) have advanced rapidly, their internal mechanisms remain opaque. Mechanistic interpretability~\citep{bereska2024mechanistic,
huben2024sparse}
seeks to bridge this gap by reverse-engineering these models and decomposing their internal computations into interpretable components. Early efforts focused on inspecting individual neurons~\citep{nguyen2019understandingneuralnetworksfeature,elhage2022superposition,bills2023language},
operating under the assumption that specific neurons would map one-to-one onto human concepts. However, this approach faced a fundamental barrier of superposition, \textit{i.e.}, neural networks represent more independent features than available neurons by encoding each feature as a linear combination of neurons~\citep{ferrando2024primerinnerworkingstransformerbased}.
Consequently, 
single neurons become difficult to interpret, as their activations represent entangled mixtures of {concepts}. 

\begin{figure*}
    \centering
    \begin{adjustbox}{width=\textwidth,center}
    \includegraphics[width=\textwidth]{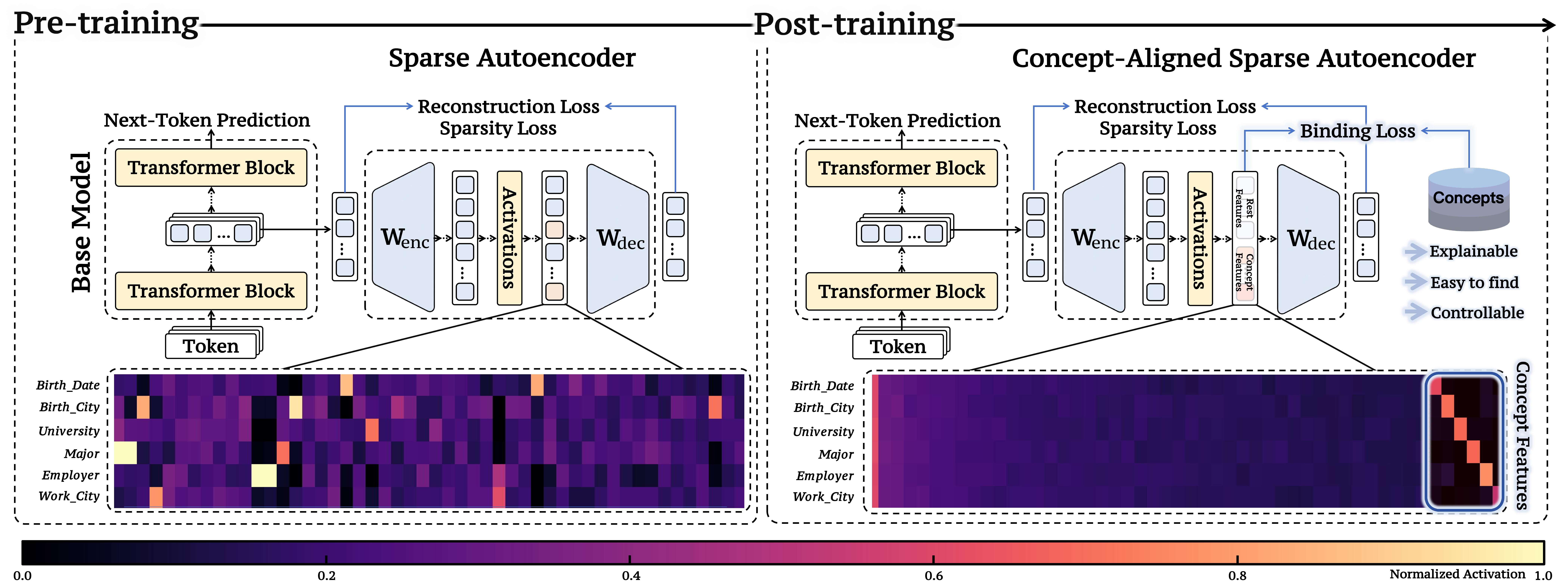}
    \end{adjustbox}
    \caption{Overview of \modelname{}. Left: An  unsupervised SAE trained post hoc on frozen LLM activations optimizes only reconstruction and sparsity, so each concept tends to be spread across multiple features, making interventions unreliable. Right: Our \modelname{} adds a supervised binding loss that maps each concept in a pre-defined ontology to a dedicated feature, yielding clean, isolated activations that are easy to find, interpret, and control.}
    \label{fig:main}
\end{figure*}

This limitation of neuron-level analysis directly motivated the development of \textit{Sparse Autoencoders (SAEs)}. The idea is to disentangle these superimposed neurons into more interpretable \textit{features}, by learning an overcomplete, sparse representation of neural activations~\citep{shu-etal-2025-survey}.
By mapping LLM's hidden states into higher-dimensional space, SAEs often learned features that are cleaner and more interpretable than individual neurons~\citep{leask2025sparse,chanin2025featurehedgingcorrelatedfeatures,yan2025visualexplorationfeaturerelationships}.
Ideally, features learned by an SAE would correspond to atomic, independent, human-interpretable concepts, so that a human can easily inspect, interpret, and steer model behavior. 
For example, in the context of relation extraction, one would expect a single SAE feature to exclusively represent the relation \textsc{Birth\_City}, such that manipulating this feature alone would precisely control the model's output regarding cities of birth. 
However, because standard SAEs are trained in an unsupervised fashion, they {\em have no explicit incentive to align their latent features with human-defined concepts}. In practice, this leads to two major challenges in interpreting the SAE feature space: 1) \textit{Feature interpretation is non-trivial}: determining which feature corresponds to a target concept is difficult. To infer a feature's semantics, practitioners often rely on constructing minimal contrast pairs~\citep{jing2025lingualensinterpretinglinguisticmechanisms,li2025saferprobingsafetyreward} or inspecting top-activating examples~\citep{cunningham2023sparseautoencodershighlyinterpretable,bereska2024mechanistic,shu-etal-2025-survey}. 2) \textit{Features remain entangled}: concepts are often fragmented across multiple features, and conversely, a single feature may correspond to multiple unrelated concepts, as shown in \cref{fig:main} (left). 
These limitations undermine downstream applications that require reliable feature-level control, such as safety steering~\citep{bereska2024mechanistic,bhattacharjee2024inferencetimecategorywisesafetysteering,ghosh2025safesteerinterpretablesafetysteering,obrien2025steeringlanguagemodelrefusal},
knowledge editing~\citep{makelov2024saes,guo2024mechanisticunlearningrobustknowledge,farrell2024applyingsparseautoencodersunlearn,zhao-etal-2025-steering,karvonen2025saebench},
and data attribution~\citep{he2024llamascopeextractingmillions,muhamed-etal-2025-decoding}.

To mitigate these issues, we take inspiration from the training pipeline of LLMs.
As illustrated in \cref{fig:main}, we view conventional SAE training as analogous to LLM pre-training: an unsupervised phase that discovers a broad latent feature space but does not guarantee alignment with human concepts. In LLMs, this misalignment is addressed by post-training steps such as instruction tuning~\citep{wei2022finetuned,zhang2025instructiontuninglargelanguage} or RLHF~\citep{ouyang2022traininglanguagemodelsfollow}.
By analogy, we propose an \textbf{SAE post-training} stage that guides the unsupervised SAE to align the SAE's feature space with a set of chosen concepts, turning it from a reconstructive tool into a reliable concept-level interface. 

Concretely, we attach an SAE to one layer of a frozen base LM and train it in two phases: first unsupervised, then supervised. After the SAE has learned to reconstruct the activations, we ``fine-tune'' the SAE with concept supervision. We reserve $|\mathcal{R}|$ dedicated concept slots (one per ontology concept), while the remaining $K$ dimensions form a free feature bank to preserve reconstruction fidelity. We then augment the training objective with additional losses to bind and isolate each concept in its corresponding feature slot. In particular, we impose: (i) a \textit{concept binding loss} that forces a one-to-one mapping between each labeled concept and a dedicated feature, (ii) a \textit{concept invariance loss} that makes each concept feature invariant to irrelevant variations and decorrelates it from the free features, and (iii) a \textit{sufficiency loss} that trains an auxiliary answer head to rely only on the concept slots for predicting concept-related information. 
Together, these objectives encourage the encoder to route concept-specific evidence into the appropriate slot rather than dispersing it across the latent space.
The result is an SAE feature space that directly corresponds to human-interpretable concepts, as shown in~\cref{fig:main} (right). In this work, we focus on the important NLP task of relation completion (RC). Accordingly, we use an ontology where concepts correspond to relation types, \emph{e.g.}, \texttt{BIRTH\_CITY}.\footnote{In principle, \modelname{} should be applicable to any type of ontology. We leave this additional study as future work.} 

Empirically, \modelname{} is more interpretable, disentangled, and controllable than a standard SAE under three RC evaluations.
\textbf{(i) Binding/generalization:} concept--slot binding becomes clean and diagonal at mid layers (\cref{fig:bind}) and transfers to unseen templates (\cref{fig:combined_binding_swap}).
\textbf{(ii) Disentanglement:} concepts concentrate onto single slots with low fragmentation and high Top-1 mass (\cref{fig:concept-frag-conc}).
\textbf{(iii) Causal control:} swap interventions exhibit a clear fidelity--strength tradeoff (\cref{tab:layer_comparison}), strong mid-layer controllability
(\cref{fig:swap-heatmap}), and predictable failure modes (\cref{tab:swap_error_type}); in 2-hop RC, post-trained \modelname{} attains $4\times$ higher swap success than the traditional SAE
(\cref{fig:2hop_swap}).

    \section{Related Work}

We review two main directions that influenced this work: \textit{Sparse Autoencoder Steering} and \textit{Concept Binding}, which address interpretable control and concept alignment, respectively. Inspired by both, we introduce \modelname{}, a lightweight, concept-aligned, and interpretable SAE framework.

\noindent\textbf{Sparse Autoencoder Steering. }
Sparse Autoencoders (SAEs) provide an interpretable interface to LLM activations by decomposing superposed, polysemantic neuron activity into sparse, overcomplete features~\citep{bricken2023monosemanticity,cunningham2023sparseautoencodershighlyinterpretable}. This representation enables \emph{SAE steering}, where intervening on specific features can causally influence model outputs toward desired behaviors or concepts~\citep{obrien2025steeringlanguagemodelrefusal,marks2025sparsefeaturecircuitsdiscovering,arad2025saesgoodsteering}. Recent work improves feature quality through training objectives and architectural choices ~\citep{rajamanoharan2024jumpingaheadimprovingreconstruction,bricken2023monosemanticity,shu-etal-2025-survey,sharkey2025openproblemsmechanisticinterpretability}. 
However, since SAE features are learned in a purely unsupervised manner, they are not guaranteed to align with a user-specified concept set: a target concept may be fragmented across multiple features, and individual features may mix unrelated signals. As a result, practical steering often still relies on manual feature identification, including contrast pairs and feature search heuristics~\citep{obrien2025steeringlanguagemodelrefusal,jing2025lingualensinterpretinglinguisticmechanisms,bayat2025steeringlargelanguagemodel,chalnev2024improvingsteeringvectorstargeting,yang2025lfsteeringlatentfeatureactivation}, 
which are limited due to the remaining poly-semanticity of the features~\citep{chanin2025featurehedgingcorrelatedfeatures,cui2025theoreticalunderstandingidentifiablesparse,minegishi2025rethinking}.
To address this, we post-train an SAE with concept supervision to learn a stable concept--feature mapping: we allocate dedicated concept slots, bind each ontology relation to a fixed slot, and make steering interventions directly targetable without post-hoc feature search. A concurrent work, G-SAE~\citep{harle2025measuring}, also conditions SAE latent representations on labeled concepts to enforce localized, monosemantic features---the fact that two independent groups arrive at this core insight strengthens the case for concept-supervised SAE training. \revision{\modelname{} goes further in several respects: (i)~we demonstrate \emph{causal controllability} via concept swaps (94.8\% success in 1-hop), a stronger test than G-SAE's feature-level monosemanticity scores (FMS); (ii)~we evaluate on \emph{compositional multi-hop reasoning}, a setting G-SAE does not address; (iii)~our value heads provide a \emph{sufficiency guarantee}---each concept slot predicts the concept's actual value, not just its label; and (iv)~we reveal a grokking-like phase transition in binding emergence, providing mechanistic insight into the alignment process.}

\noindent\textbf{Concept Binding. } Posterior Regularization (PR) ~\citep{JMLR:v11:ganchev10a} and Logic Rule Encoding (LRE) ~\citep{hu-etal-2016-harnessing} are traditional frameworks adopted to bind human-defined concepts to neural models by imposing soft constraints on posterior distribution, or integrating logic rules into the learning objectives. 
Prior work has applied PR to reading comprehension by enforcing linguistic concept-level constraints~\citep{zhou2019robustreadingcomprehensionlinguistic}, 
and to question answering by mapping event triggers to conceptual constraints~\citep{lu2023eventknowledgeincorporationposterior}. LRE works~\citep{hu-etal-2016-harnessing, fischer2019dl2, yang2023injectinglogicalconstraintsneural} reconstruct the training objective by combining the task loss with a rule loss, thereby binding logical concepts to model predictions via parameter updates. Although LRE addresses the soft-constraint issue of PR, it remains a black-box mechanism that cannot be used to interpret or control specific internal representations of the model. 
Concept Bottleneck Models~\citep{koh2020conceptbottleneckmodels}, originally proposed for image classification, map intermediate representations to human-readable concepts via a supervised concept-mapping loss.
However, such heavy-weight intervention into the base model architecture is not easily scalable to large-scale models.
To address these gaps, we propose a lightweight and interpretable framework, \modelname{}, which binds human-readable concepts to the intermediate representations of a frozen base model.

\revision{
\paragraph{Disentangled Representation Learning.}
\modelname{} draws on principles from the disentanglement literature.
$\beta$-VAE~\citep{higgins2017betavae} and
FactorVAE~\citep{kim2018disentangling} use modified ELBO
objectives to encourage factorial latent distributions;
our alignment loss serves an analogous purpose but with
\emph{explicit} concept supervision rather than unsupervised
statistical independence.
We adopt the Mutual Information Gap
(MIG)~\citep{chen2018isolating} as a quantitative
disentanglement metric (\S\ref{app:mig}).
Concept Bottleneck Models~\citep{koh2020conceptbottleneckmodels} create
interpretable intermediate representations aligned with
human-readable concepts; \modelname{} can be viewed as a
post-hoc concept bottleneck applied to frozen model
activations, without modifying the base model architecture.
A key distinction from classical disentanglement is that
standard methods assume \emph{statistically independent}
factors of variation---an assumption that often fails for
language (e.g., employer and work\_city are correlated).
Our experiments confirm this: a standard SAE trained with only
reconstruction and sparsity achieves MIG of exactly 0.000
for all 1-hop concepts, because without explicit supervision,
no single feature has any incentive to exclusively capture a
specific concept. \modelname{}'s supervised binding addresses
this directly.
}

    \section{Method}
\label{sec:approach_overview}

We construct an explicit, verifiable, and controllable interface over a frozen LLM's activations via a Concept-Aligned Sparse Autoencoder (\modelname{}). This interface provides evidence that the model encodes concept-relevant information in a linearly accessible form, though the one-to-one slot mapping is a product of the supervised training objective rather than necessarily reflecting the model's native representational structure.

\noindent\textbf{Terminology. }
In this work, we focus on the task of relation completion (RC), where facts are represented as triples \((e_1,r,e_2)\): the relation type \(r\) links entity mentions \(e_1\) and \(e_2\) (\textit{e.g.}, \(\big(\text{\textit{``Marie Curie''}}, \textsc{born\_in}, \text{\textit{``Warsaw''}}\big)\)).
In this paper, RC appears as \emph{two tasks}. 
For 1-hop, given \((e_1,r)\), the model predicts the missing object \(e_2\); for instance, \textit{``What is Alice's \texttt{birth date}?''} corresponds to a single relation query (\cref{subsec:task_1hop}). 
For 2-hop, given \((e_1,r_1,r_2)\), the model predicts \(e_3\) by composing two relations through an intermediate entity \(e_2\), \textit{i.e.}, \(e_1\xrightarrow{r_1}e_2\xrightarrow{r_2}e_3\); \textit{e.g.}, \textit{``Who is the \texttt{classmate} of the \texttt{mentor} of Barbara?''} (\cref{subsec:task_2hop}).
Here we instantiate {concepts} as relation types from a domain {ontology}, a finite inventory of semantic links between entities (\textit{e.g.}, \textsc{born\_in}, \textsc{friend\_of}). We focus on relations because RC is a core NLP task with high-impact closed-domain applications (\textit{e.g.}, biomedicine~\citep{bionlp-ws-2025-1}, finance~\citep{grishman-sundheim-1996-message}), where the target relations are defined by an ontology; nevertheless, our framework is not tied to relations and could bind other ontology concepts (\textit{e.g.}, entities or attributes) to supervised SAE slots.

\noindent\textbf{Concept Binding in Activation Space. }
Given a frozen LLM $\mathcal{M}$, we extract an activation $h \in \mathbb{R}^d$ from layer $\ell$ and learn a sparse representation $z = E(h) \in \mathbb{R}^{|\mathcal{R}|+K}$.
We partition $z = [z_{\text{rest}}; z_{\text{concept}}]$, where
$z_{\text{concept}} \in \mathbb{R}^{|\mathcal{R}|}$ are the
supervised concept slots and
$z_{\text{rest}} \in \mathbb{R}^{K}$ are the free features
($|\mathcal{R}| \ll K$).
\revision{
We define the average activation of concept $k$ on slot $j$ as:
$	\bar{a}_{k,j} = \mathbb{E}_{x \sim \mathcal{D}_k}
	\left[ z_j^{(k)}(x) \right]$,
where $\mathcal{D}_k$ denotes the subset of inputs labeled
with concept $k$, and $z_j^{(k)}(x)$ is the activation of
slot $j$ for input $x$ with concept label $k$. The supervised
slots are indexed as
$j \in \{K{+}1, \ldots, K{+}|\mathcal{R}|\}$.
}
We supervise $z_{\text{concept}}$ so that each ontology concept occupies its own dedicated slot: for an input containing concept $c$, its slot should activate while other concept slots are suppressed. This makes concept--slot activations a direct, verifiable readout of the concept in use. We keep $z_{\text{rest}}$ ``free'' to absorb residual linguistic variation and keep the base model's statistical structure.

\noindent\textbf{A Verifiable Interface. }
Unlike a traditional SAE, which is primarily diagnostic, our approach constructs an operational interface that can be validated by intervention rather than relying on minimal comparison pairs. Specifically, (i) \emph{verification}: we can check whether the model is using a particular relation by observing whether the corresponding concept slot activates; and (ii) \emph{control}: we can causally steer the computation by manually activating or suppressing that slot, directly influencing the model’s downstream prediction.

\noindent\textbf{Model Training. }
Although \modelname{} is LLM-agnostic, we evaluate it on GPT-2.\footnote{We choose GPT-2 for controlled analysis under compute limits, enabling layer-wise probing, ablation/steering analysis.} We fine-tune the base model on our RC training data (\cref{app:training}), extract layer activations at the final question token across all layers, and train \modelname{} with a task-dependent free-feature bank of size $K$ and $|\mathcal{R}|$ concept slots using a two-stage training (\cref{subsection:implementation_alignsae,app:sae}).

\noindent\textbf{SAE Pre-Training and Post-Training. }
Directly optimizing the full objective from scratch can produce unstable binding (\cref{app:stage1_ablation}), so we adopt a two-phase curriculum that parallels LLM pre-/post-training (\cref{fig:main}). In the \emph{pre-training} phase, the SAE is trained on reconstruction and sparsity, allowing the decoder to form a high-capacity dictionary.
In subsequent \emph{post-training}, we strengthen the binding and value losses, then activate the orthogonality penalty. This reshapes the latent space so supervised slots become clean, disentangled carriers of atomic concepts, while remaining decoupled from the free feature bank.
This curriculum retains the benefits of joint optimization (concept slots that are simultaneously interpretable and task-predictive) while avoiding the degenerate minima that arise when strong supervision is applied before the underlying representation has stabilized (\cref{app:sae}).

\noindent\textbf{Objectives. }
\label{para:objectives}
We train the encoder, decoder, and value head jointly. Our binding objective augments the standard SAE loss ($\mathcal{L}_{\text{SAE}}$) with (i) a supervised loss that assigns each relation to a dedicated concept slot ($\mathcal{L}_{\text{align}}$), (ii) a $z_{\text{concept}}$--$z_{\text{rest}}$ decorrelation penalty to reduce concept leakage into $z_{\text{rest}}$ ($\mathcal{L}_{\perp}$), and (iii) an auxiliary value loss that encourages the concept slots to support answer prediction ($\mathcal{L}_{\text{val}}$)~\citep{agarwal2021neural,hewitt2023backpack,sun2024conceptbottleneck}, where $y_{\text{ans}}$ is the ground-truth answer:
\begin{equation}
\begin{aligned}
& \mathcal{L}_{\text{SAE}} = \lambda_{\text{rec}}\|h-\hat{h}\|_2^2 + \lambda_{\text{sp}}\|z\|_1, \\
& \mathcal{L}_{\text{align}} = \mathrm{CE}\big(\mathrm{softmax}(z_{\text{concept}}), y_{\text{rel}}\big), \\
& \mathcal{L}_{\perp} = \|\mathrm{corr}(z_{\text{concept}}, z_{\text{rest}})\|_F^{2}, \\
& \mathcal{L}_{\text{val}} = \mathrm{CE}\big(\mathrm{softmax}(V(z_{\text{concept}})), y_{\text{ans}}\big), \\
& \mathcal{L} = \mathcal{L}_{\text{SAE}} + \lambda_{\text{align}}\mathcal{L}_{\text{align}} + \lambda_{\perp}\mathcal{L}_{\perp} + \lambda_{\text{val}}\mathcal{L}_{\text{val}}.
\end{aligned}
\end{equation}
See \cref{app:loss_function} for full definitions and hyperparameters. We also assess each term through ablations in \cref{app:ablation}.

Overall, our method turns a frozen LLM’s conceptual evidence into an explicit, verifiable, and controllable interface by (i) training an SAE on a single intermediate layer, (ii) allocating one-to-one ontology-aligned concept slots, and (iii) using a two-stage curriculum that yields disentangled, slot-addressable features. Because only the lightweight SAE is trained, the base model remains fixed, enabling efficient diagnosis (slot verification) and causal control (slot interventions).
    \section{Implementation}
\label{sec:implementation}

We attach the SAE to a frozen base LLM (\textit{e.g.}, GPT\textendash2; our approach is not limited to a specific model) by extracting a representation $h\in\mathbb{R}^d$ from an intermediate layer $\ell$. Concretely, given an input biography question $x$, the model produces token-level hidden states at each layer, and we pool the activations from layer $\ell$ 
(\textit{e.g.}, mean pooling)
to form $h$. The base LLM is kept frozen, preserving its linguistic competence while placing all supervision on the light-weight SAE interface (\cref{app:activation_collection}).

\subsection{Concept-Aligned Sparse Autoencoder}
\label{subsection:implementation_alignsae}
We use a supervised SAE that exposes an interpretable control surface aligned with $\mathcal{R}$, while delegating residual variance to a large bank of free features. The encoder $E:\mathbb{R}^{d}\to\mathbb{R}^{|\mathcal{R}|+K}$ maps $h$ to a sparse code $z=\mathrm{ReLU}(W_e h+b_e)$, partitioned as $z=[z_{\text{concept}};z_{\text{rest}}]$ with $z_{\text{rest}}\in\mathbb{R}^{K}$ (unsupervised free features) and $z_{\text{concept}}\in\mathbb{R}^{|\mathcal{R}|}$ (supervised concept slots). The decoder $D:\mathbb{R}^{|\mathcal{R}|+K}\to\mathbb{R}^{d}$ reconstructs $\hat h=W_d z+b_d$. We set $K{=}10{,}000$ (1-hop) and $K{=}100{,}000$ (2-hop) to add capacity without burdening the $|\mathcal{R}|$ interpretable slots, read out concept evidence from $z_{\text{concept}}$ at the output token, and post-train the SAE with the loss in \cref{para:objectives}.

\subsection{Tasks and Datasets}
\label{subsec:tasks}

We evaluate our approach \modelname{} on two controlled benchmarks designed to test both direct attribute retrieval and multi-step logical derivation: (i) \textit{biography factual recall} (1-hop) and (ii) \textit{multi-step compositional reasoning} (2-hop).

\subsubsection{Factual Recall (1-hop)}
\label{subsec:task_1hop}
We validate our approach on a biography QA task over a fixed ontology. Let $\mathcal{R}$ be a fixed ontology of $|\mathcal{R}|=6$ (\cref{tab:qa_templates}; \cref{app:dataset}). For a person $p$ and relation $r\in\mathcal{R}$, a canonical table provides the gold value
$y^\star=g(p, r)$ (\textit{e.g.}, \emph{Wesleyan University} for \textsc{university}).
Each input $x$ is a natural-language question that mentions $p$ and implicitly targets a single relation
$r^\star\in\mathcal{R}$; the model must generate the corresponding value $y^\star=g(p, r^\star)$.

We adopt the synthetic biography dataset of~\citet{AL2023-knowledge1} with minor modifications to better separate \emph{semantic binding} from \emph{template memorization}. We generate 1{,}000 person profiles (five biography variants each) and instantiate questions from paraphrase templates that preserve the same underlying relation. 
To assess robustness to surface form, we use a disjoint template split: we train on all profiles using two templates and evaluate \textsc{Unseen-Template} generalization on two held-out templates (\cref{tab:qa_templates}; \cref{app:dataset}).
The held-out templates use different lexical triggers (\textit{e.g.,} \emph{born} vs.\ \emph{birth city}, \emph{alma mater} vs.\ \emph{university}) to reduce n-gram overlap with training prompts. 
We also explored \emph{LLM-generated questions} by few-shot prompting Claude~3.5~Sonnet on 1-hop RC queries $(e_1,r)$ to generate instance-specific paraphrases, but excluded them since many are too indirect for base GPT-2 (details in \cref{fig:ood_prompt}; \cref{app:llm_templates}).

\subsubsection{Compositional Reasoning (2-hop)}
\label{subsec:task_2hop}
\begin{figure}[t]
    \centering
    \includegraphics[width=0.5\linewidth]{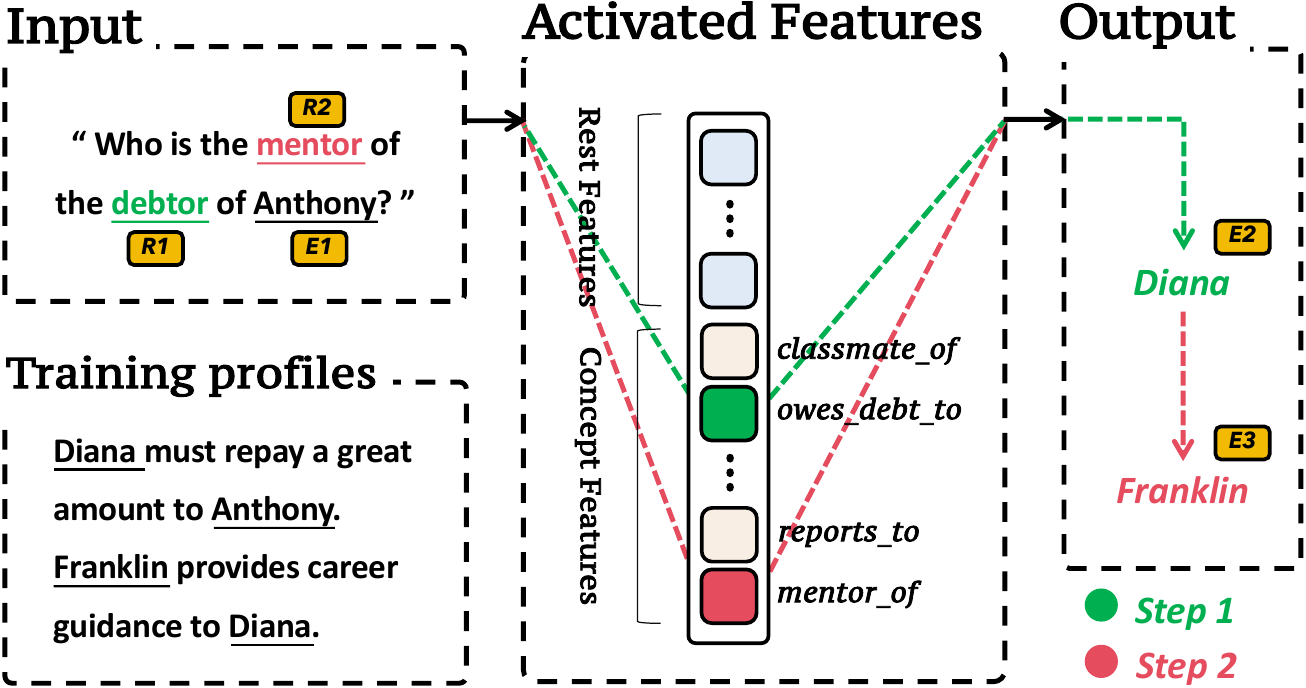}
    \caption{{2-hop step-wise alignment.} With \((e_1,r_1,r_2)\), the model predicts \((e_2,e_3)\); \(r_1\) activates at \(e_2\) and \(r_2\) at \(e_3\), with remaining capacity in unsupervised features.}
    \label{fig:2hop_overview}
\end{figure}
We further evaluate \modelname{} on a \emph{two-hop} compositional reasoning task~\citep{du-etal-2025-reason}. We use 2-hop as a compositional concept test for \modelname{}. Unlike 1-hop, it decomposes prediction into two concept steps with an intermediate entity $e_2$, so concept--slot binding can be \emph{verified} and \emph{intervened} at the step where each concept is actually used, rather than inferred only from final-answer accuracy. This decomposition also enables targeted causal tests (\textit{e.g.}, swapping the $r_2$ at the second step), making controllability more diagnostic.
Each example provides a start entity $e_1$ and relations $(r_1,r_2)$; the model must infer
$e_1 \xrightarrow{r_1} e_2 \xrightarrow{r_2} e_3$ (\emph{i.e.}, $e_2{=}r_1(e_1)$, $e_3{=}r_2(e_2)$).

Inputs use templates like \textit{``Who is the $r_2$ of the $r_1$ of $e_1$?''} with brief profile sentences stating the two hop facts (\cref{fig:twohop_example_box}; \cref{app:twohop_format}).
The dataset instantiates an ontology of 20 relations over 60 entities (see \cref{app:twohop_format}).
Following~\cref{fig:2hop_overview}, we train the model to output “$e_2;e_3$” and apply {step-wise} supervision: the $r_1$ slot should activate when generating $e_2$, and the $r_2$ slot when generating $e_3$.

\subsection{Metrics}
\label{subsec:metrics}
To measure concept–slot alignment, we build a relation–slot confusion matrix $C$ where $C[r,j]$ is the normalized fraction of examples whose top-activated slot at the answer token is $j$ for gold relation $r$, and report the \emph{diagonal accuracy} $\mathrm{Diag}:=\frac{1}{|\mathcal{R}|}\sum_{r\in\mathcal{R}} C[r,r]$ ($1$ for perfect one-to-one binding, $\approx 1/|\mathcal{R}|$ at chance; see \cref{app:metrics}).

\subsection{Controllability (Swap Test)}
\label{subsec:swap_test}
To evaluate \emph{causal} controllability, we perform a \emph{swap test}: given a query whose gold relation is $r^\star$, we intervene by injecting the decoded direction of an alternative relation slot $j\neq r^\star$ with strength $\alpha>0$.
Formally, letting $h_{\ell,t}\in\mathbb{R}^d$ denote the residual stream at layer $\ell$ and token position $t$, and $W_{\text{dec}}\in\mathbb{R}^{d\times K}$ denote the SAE decoder with reconstruction $\hat h_{\ell,t}=W_{\text{dec}} z_{\ell,t}$, the decoded basis direction for slot $j$ is $v_j := W_{\text{dec}} e_j$ and the intervention is
$h'_{\ell,t} \;=\; h_{\ell,t} + \alpha\, v_j \;=\; h_{\ell,t} + W_{\text{dec}}(\alpha e_j)$.
We count a swap as successful if the model’s top-1 generated answer matches the gold value under the intervened relation, \emph{i.e.}, $g(p,j)$, rather than the original target $g(p,r^\star)$ (full definitions in~\cref{app:metrics}).

\section{Results: Factual Recall (1-hop)}
\label{sec:experimental_setup}

\begin{wrapfigure}{r}{0.4\linewidth}
    \centering
    \includegraphics[width=\linewidth]{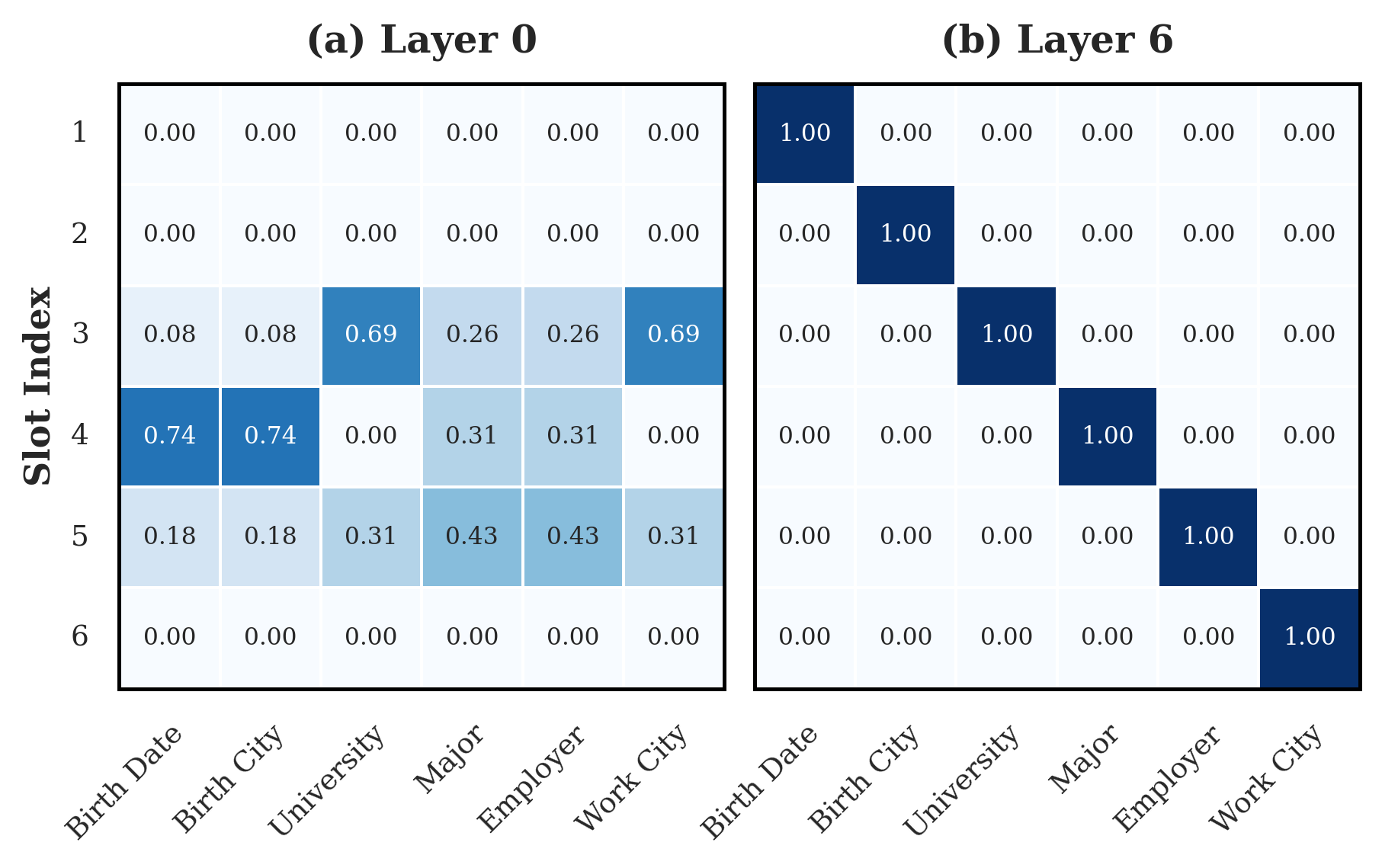}
    \caption{Concept–slot binding at a shallow layer (a) versus a mid layer (b). At layer 0, supervision for each relation is dispersed across multiple slots, whereas at layer 6 the SAE learns a perfect one-to-one binding.}
    \label{fig:bind}
\end{wrapfigure}
We first evaluate the proposed interface across transformer layers, data (Train, Test\textendash Unseen), and controllability settings. Metrics follow~\cref{subsec:metrics}.

\subsection{Layer-Wise Performance}



The contrast between layer~0 and layer~6 illustrates how concept alignment emerges (\cref{fig:bind}). At layer~0, the relation\textendash slot confusion matrix is diffuse with substantial off-diagonal mass and overlapping activations, indicating entangled features that fail to isolate ontology relations. At layer~6, the same matrix collapses to a sharp diagonal with negligible off-diagonal mass, yielding a stable permutation that assigns each relation to a single slot. 

\begin{figure}[t]
    \centering
    \includegraphics[width=0.48\linewidth]{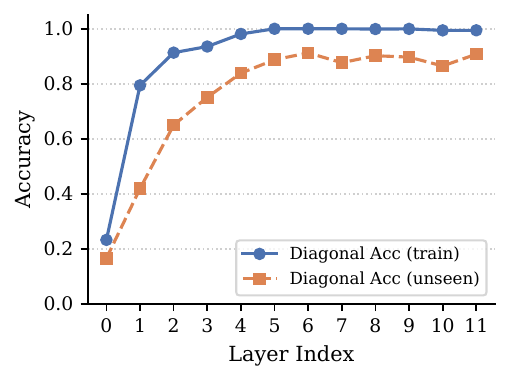}
    \hfill
    \includegraphics[width=0.48\linewidth]{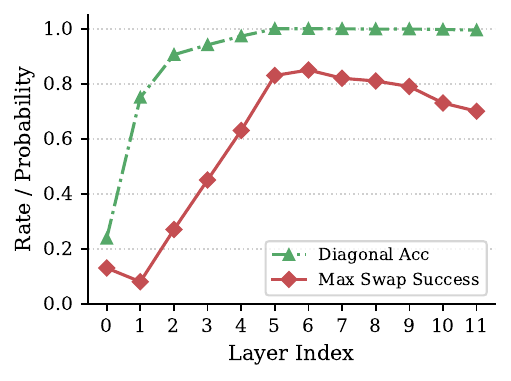}
    \caption{Binding generalization vs.\ causal control across layers. Left: diagonal accuracy on train (blue) and on unseen templates (orange). Right: diagonal accuracy and maximum swap success over $\alpha$, both on held-out test questions.}
    \label{fig:combined_binding_swap}
\end{figure}

\cref{tab:layer_comparison} and \cref{fig:combined_binding_swap} reveal a clear mid-layer sweet spot. Performance peaks around Layer~6, where binding is essentially one-to-one (\textit{Diagonal Acc (test)}$=1.00$), swap controllability is highest, and paraphrase generalization remains strong (\textit{Diagonal Acc (unseen)}$=0.912$). In contrast, early layers are poorly aligned (Layer~0, \textit{Diagonal Acc (test)}$=0.238$), with diffuse off-diagonal mass (\cref{fig:bind}). Although Layer~6 incurs higher reconstruction error, it yields a large gain in controllability (\textit{swap success} $+0.81$), suggesting that clean semantic binding emerges in mid layers with richer contextual representations, while early \mbox{embedding-dominated} layers and deeper, more compressed layers make a stable slot interface harder to preserve.

\subsection{Concept–Feature Alignment}
\label{subsec:layerwise-frag}

To quantify how cleanly each concept is represented at different depths, we compare a traditional SAE trained purely unsupervised with \modelname{}, which adds concept-supervision as a post-training signal. Let $z_i \in \mathbb{R}^{|\mathcal{R}|+K}$ be the sparse code for example $i$, and let $c(i)$ be its concept label (\textit{e.g.}, one of six ontological relations). We first define the \emph{average activation} of concept $c$ on feature $k\in\{1,\dots,|\mathcal{R}|\}$ as $A_{c,k} = \mathbb{E}_{i : c(i) = c}\big[z^{(i)}_{\text{concept},k}\big],$ and normalize over features to obtain a concept–feature distribution $B_{c,k} = \frac{A_{c,k}}{\sum_{k'} A_{c,k'} + \epsilon}.$ From $B_{c,\cdot}$ we derive two summary metrics for each concept $c$:

\begin{wraptable}[12]{r}{0.55\linewidth} 
    \centering
    \setlength{\tabcolsep}{3pt}
    \small 
    \begin{tabular}{lccc}
    \toprule
    \textbf{Metric} & \textbf{Layer 0} & \textbf{Layer 6} & \textbf{$\Delta$ (L6 -- L0)} \\
    \midrule
    Diagonal Acc (test) $\uparrow$ & 0.238 & \textbf{1.000} &
      \cellcolor{green!15}{\textcolor{green!60!black}{\textbf{$\uparrow\ 0.76$}}} \\
    Swap Success $\uparrow$      & 0.040 & \textbf{0.847} &
      \cellcolor{green!20}{\textcolor{green!60!black}{\textbf{$\uparrow\ 0.81$}}} \\
    Diagonal Acc (train) $\uparrow$   & 0.232 & \textbf{1.000} &
      \cellcolor{green!15}{\textcolor{green!60!black}{\textbf{$\uparrow\ 0.77$}}} \\
    Diagonal Acc (unseen) $\uparrow$   & 0.165 & \textbf{0.912} &
      \cellcolor{green!15}{\textcolor{green!60!black}{\textbf{$\uparrow\ 0.75$}}} \\
    Recon MSE $\downarrow$        & $6.53\cdot 10^{-5}$ & $7.42\cdot 10^{-2}$ &
      \cellcolor{red!15}{\textcolor{red!70!black}{\textbf{$\uparrow\ \approx 1.1\text{k}\times$}}} \\
    \bottomrule
    \end{tabular}
    \caption{Layer 0 vs.\ Layer 6 performance comparison.}
    \label{tab:layer_comparison}
    \vspace{-50pt} 
\end{wraptable}

\noindent\textbf{Effective number of features (EffFeat). }
We measure how many features are effectively used to represent a concept via the entropy
of $B_{c,\cdot}$:
$    \text{EffFeat}(c) = \exp\Big(- \sum_{k} B_{c,k} \log B_{c,k}\Big),$
where smaller values indicate that a concept is concentrated on fewer features
(\emph{i.e.}, lower fragmentation).

\noindent\textbf{Top-1 concentration (Top1C). }
We also track how much of a concept’s mass is captured by its single most responsive
feature:
$\text{Top1C}(c) = \max_{k} B_{c,k},$
where larger values indicate that one feature dominates the representation of
concept $c$.

\begin{figure}[ht]
    \centering
    \begin{subfigure}{0.48\linewidth}
        \centering
        \includegraphics[width=\linewidth]{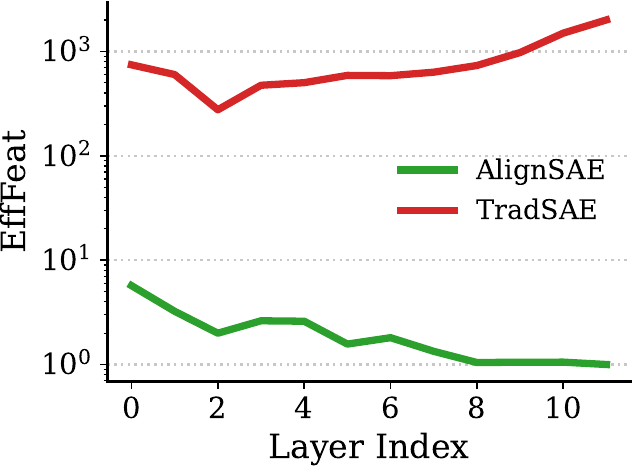}
        \caption{Concept fragmentation ($\downarrow$)}
        \label{fig:concept-frag-eff}
    \end{subfigure}
    \hfill
    \begin{subfigure}{0.48\linewidth}
        \centering
        \includegraphics[width=\linewidth]{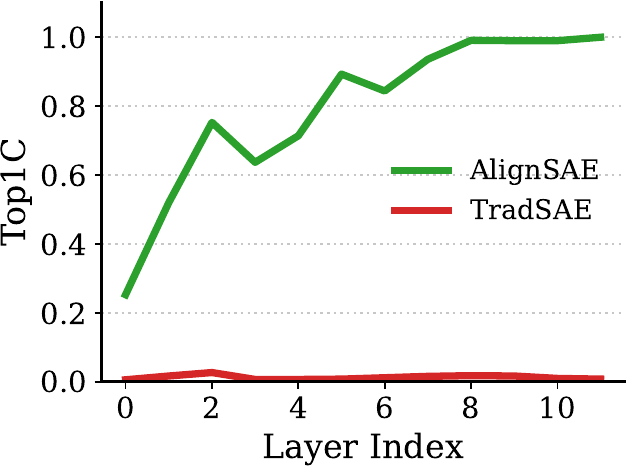}
        \caption{Concept concentration ($\uparrow$)}
        \label{fig:concept-frag-top1}
    \end{subfigure}
    \caption{Layer-wise fragmentation ($\downarrow$) and concentration ($\uparrow$) for \modelname{} and a traditional SAE.}
    \label{fig:concept-frag-conc}
\end{figure}

\cref{fig:concept-frag-conc} reports layer-wise averages of these metrics over six concepts for \modelname{} and a traditional SAE. The traditional SAE (pre-training only) remains highly fragmented across all layers: $\text{EffFeat}$ spans hundreds to thousands of features per concept and $\text{Top1C}\approx0$, suggesting that concepts are not consistently localized to a single feature. In contrast, \modelname{} sharply reduces fragmentation (EffFeat $\approx 1$) and increases concentration via post-training supervision, yielding compact representations from mid layers onward and near one-to-one bindings in deeper layers. Overall, concept-level post-training converts a diffuse many-to-many feature space into a compact, interpretable interface with directly addressable concepts.

Detailed disentanglement and robustness analyses are deferred to \cref{app:mig,app:latent_vis,app:side_effects}.

\subsection{Swap Controllability}
\label{subsec:swap_result}
\begin{wrapfigure}[14]{r}{0.45\linewidth}
  \centering
  \includegraphics[width=\linewidth]{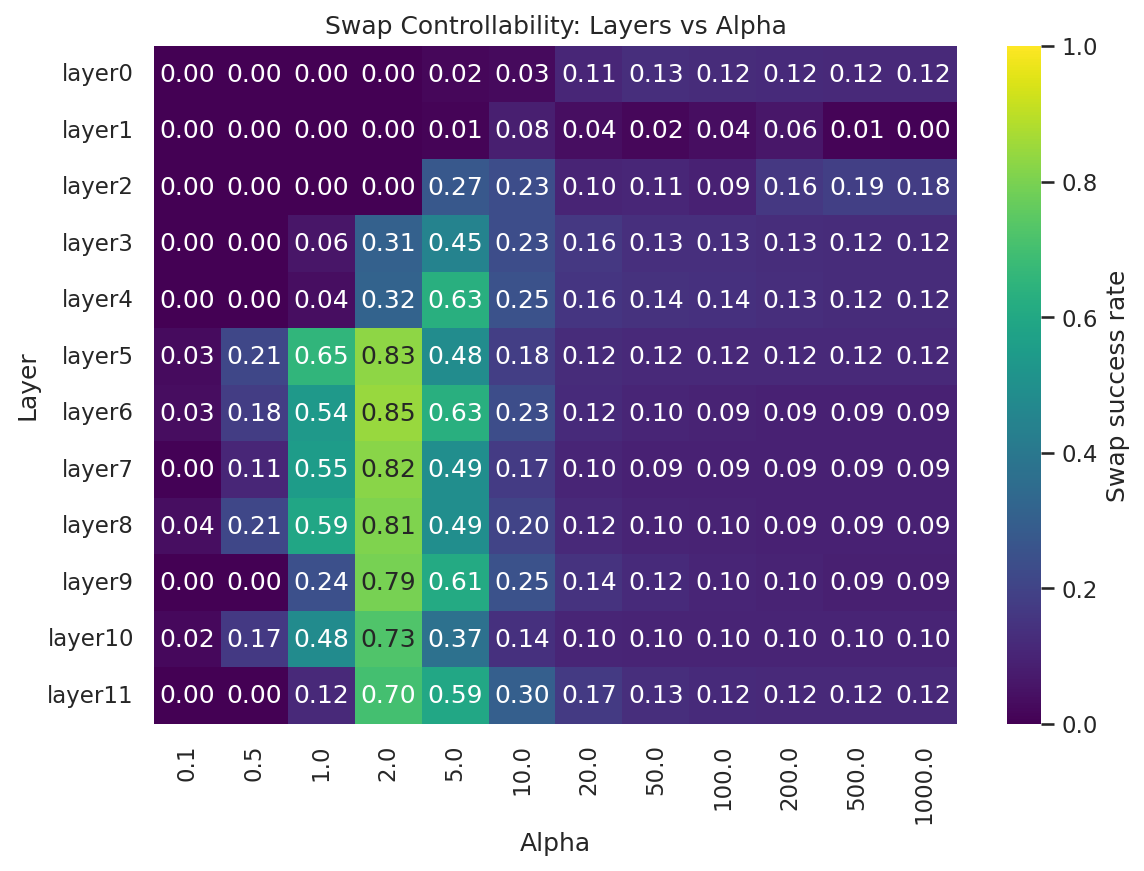}
  \caption{Swap controllability across layers and amplification $\alpha$. Mid layers remain robust at $\alpha\!\approx\!2$.}
  \label{fig:swap-heatmap}
\end{wrapfigure}

We probe whether concept slots behave as usable control knobs by measuring swap success under intervention (see~\cref{subsec:swap_test}).
\cref{fig:swap-heatmap} reports average swap success rate (average over 1{,}000 examples) across layers and amplification strengths: moderate amplification ($\alpha\!\approx\!2$) reliably switches the answer type at mid layers (5--8).
At layer~6, success rises from $0.54$ at $\alpha{=}1.0$ to $0.85$ at $\alpha{=}2.0$, but drops to $0.23$ at $\alpha{=}10.0$, indicating that over-amplification destabilizes the intervention and revealing a narrow regime where control is effective and predictable.

For example, for a \textsc{birth\_date} question, the model originally answers \textit{``24, March, 1964''}; amplifying the \textsc{university} slot with $\alpha{=}2$ switches the output to \emph{Wesleyan University}, showing that slots are causal control handles rather than merely diagnostic (more examples in \cref{tab:swap_examples_correct} \& \cref{fig:swap};~\cref{app:swap_examples_correct}).
We further ablate the post-training objectives; results in~\cref{app:ablation} show each component is necessary for robust controllability.
Overall, early layers are too local and deeper layers too compressed; mid-layer post-training (\textit{e.g.}, Layer~6) yields near-perfect binding and reliable swaps at $\alpha\!\approx\!2$, the practical regime for controllable concept access.
Detailed side-effects evaluation (perplexity under correct/wrong-slot steering and LLM-judge free-form quality) is reported in \cref{app:side_effects}.

\subsection{Swap Error Analysis}
\label{subsec:error_analysis}

\definecolor{headerblue}{RGB}{240, 245, 255}
\definecolor{mwpgreen}{RGB}{46, 125, 50}
\definecolor{mwpred}{RGB}{198, 40, 40}

\begin{wraptable}[14]{r}{0.6\linewidth}
    \centering
    \small
    \setlength{\tabcolsep}{2pt}
    \begin{tabular}{lcccccc}
    \toprule
    \multirow{2}{*}{\textbf{Target swap}} &
    \multicolumn{3}{c}{\textbf{$\alpha{=}2$ (15\% Err)}} &
    \multicolumn{3}{c}{\textbf{$\alpha{=}10$ (77\% Err)}} \\
    \cmidrule(lr){2-4}\cmidrule(lr){5-7}
    & \textbf{S} & \textbf{D} & \textbf{S\%} & \textbf{S} & \textbf{D} & \textbf{S\%} \\
    \midrule
    birth\_city  & 36 & 20 & 64.3 & 91  & 50 & 64.5 \\
    birth\_date  & 20 & 0  & 100.0& 139 & 0  & 100.0 \\
    employer     & 8  & 1  & 88.9 & 134 & 3  & 97.8 \\
    major        & 1  & 0  & 100.0& 77  & 51 & 60.2 \\
    university   & 42 & 10 & 80.8 & 101 & 1  & 99.0 \\
    work\_city   & 9  & 7  & 56.2 & 94  & 25 & 79.0 \\
    \midrule
    \textbf{Overall} & \textbf{116} & \textbf{38} & \textbf{75.3} & \textbf{636} & \textbf{130} & \textbf{83.0} \\
    \bottomrule
    \end{tabular}
    \caption{Category retention on \emph{failed} swaps at Layer 6. ``Same'' (S) means correct semantic class; ``Diff'' (D) is outside.}
    \label{tab:swap_error_type}
\end{wraptable}
Even when swap steering fails to hit the \emph{exact} gold entity, it often preserves the
\emph{answer category} of the intended swapped concept.
\cref{tab:swap_error_type} reports \emph{Category Preservation}, computed \emph{only over failed swaps}. Under moderate amplification ($\alpha{=}2$), $75.3\%$ of failures remain in the
correct category; under strong amplification ($\alpha{=}10$), this rises to $83.0\%$, suggesting
that larger interventions more reliably move the model onto the right target type even
as exact entity selection degrades.

For example, in \cref{tab:swap_failure_example}, swapping \textsc{university}$\rightarrow$\textsc{major} at $\alpha{=}10$ yields a \textcolor{mwpgreen}{\textit{\textbf{type}} \cmark} but \textcolor{mwpred}{\textit{\textbf{entity}} \xmark} answer: \emph{Geography} \mbox{instead} of \emph{Physical Therapy}.
In this controlled setting, our strict swap metric may understate controllability: even without exact entity matches, interventions often enforce the target concept.

\begin{table}[t]
    \centering
    \small
    \renewcommand{\arraystretch}{1.4} 
    \begin{tabularx}{\linewidth}{@{} p{1.5cm} X @{}}
    \toprule
    \rowcolor{headerblue}
    \textbf{Swap} & \textbf{Q:} Where did Jesse Kian Tate go to college? \\
    \rowcolor{headerblue}
    & \textsc{univ} $\rightarrow$ \textbf{Swap:} \textsc{major} \\
    \midrule
    \textbf{Outputs} & \textbf{Base:} Rochester Inst. of Tech. \\
    & \textbf{Gold:} Physical Therapy \\
    & \textbf{Gen:} \colorbox{gray!15}{\textbf{Geography}} \hfill 
      {\small 
      \textbf{[} \textcolor{mwpgreen}{Type \faCheck} \quad \textcolor{mwpred}{Entity \faTimes} \textbf{]}
      } \\
    \bottomrule
    \end{tabularx}
    \caption{A failure case: steering ($\alpha = 10$) successfully flips the semantic category (type) but fails to retrieve the specific ground-truth entity.}
    \label{tab:swap_failure_example}
\end{table}

    \section{Ablation Study}
\label{app:ablation}

\subsection{Effect of Training Objectives}
\label{app:abl_loss}
We ablate three post-training components of \modelname{}: (i) {binding/alignment} (\(\mathcal{L}_{\text{align}}\)), (ii) {orthogonality} (\(\mathcal{L}_{\perp}\)), and (iii) {value/sufficiency} (\(\mathcal{L}_{\text{val}}\)), while keeping the underlying task and backbone fixed.

\begin{table*}[t]
\centering
\renewcommand{\arraystretch}{1.15}
\setlength{\tabcolsep}{5pt}
\begin{tabular}{lccccc}
\toprule
\textbf{Config} &
\makecell{\textbf{Train}\\\textbf{Diag.\ Acc}} &
\makecell{\textbf{Test-Unseen}\\\textbf{Diag.\ Acc}} &
\makecell{\textbf{Train}\\\textbf{Acc}} &
\makecell{\textbf{Test-Unseen}\\\textbf{Acc}} &
\makecell{\textbf{Swap}\\\textbf{Success Rate}} \\
\midrule
\textbf{joint}    & \textbf{1.000} & \textbf{0.912} & \textbf{1.000} & \textbf{0.809} & \textbf{0.847} \\
no\_align         & 0.167 & 0.167 & \textbf{1.000} & \textbf{0.809} & 0.053 \\
no\_ortho         & 0.711 & 0.689 & \textbf{1.000} & \textbf{0.809} & 0.046 \\
no\_value         & 0.177 & 0.172 & \textbf{1.000} & \textbf{0.809} & 0.720 \\
\bottomrule
\end{tabular}
\caption{\textbf{Ablations of \modelname{} post-training.}
\textbf{Best} configuration is \textbf{joint}, which achieves strong OOD binding and high swap success simultaneously.
Swap success is reported at each configuration's best-performing intervention strength
(\(\alpha\): joint=2, no\_align=0.1, no\_ortho=10, no\_value=100). Chance diagonal accuracy for 6 relations is \(1/6\approx 0.167\).}
\label{tab:ablation}
\end{table*}

\paragraph{Results.}
As shown in \cref{tab:ablation}, \textbf{binding is necessary for alignment:} removing \(\mathcal{L}_{\text{align}}\) collapses diagonal accuracy to chance (0.167 Test-Unseen) and nearly eliminates swap control (0.053).
\textbf{Orthogonality is necessary for control:} without \(\mathcal{L}_{\perp}\), binding remains high (0.689 Test-Unseen) but swaps fail (0.046), indicating leakage that breaks causal interventions.
\textbf{Value improves interface quality:} removing \(\mathcal{L}_{\text{val}}\) yields near-chance diagonal accuracy (0.172 Test-Unseen) yet can still achieve moderate swap success (0.72) at very large \(\alpha\), whereas the full model achieves both strong Test-Unseen diagonal accuracy (0.891) and high swap success (0.847) at a small \(\alpha=2\).

\subsection{Effect of Stage-1 Pretraining}
\label{app:stage1_ablation}

We ablate the unsupervised Stage~1 and train \modelname{} \emph{from scratch} with the same post-training objective (\emph{No Stage~1}), keeping all other settings fixed.

\begin{table}[t]
\centering
\footnotesize
\renewcommand{\arraystretch}{1.10}
\setlength{\tabcolsep}{5pt}
\begin{tabular}{lcccc}
\toprule
\textbf{Metric} & \textbf{No Stage 1} & \textbf{With Stage 1} & \textbf{$\Delta$ (With--No)} & \textbf{\%$\Delta$} \\
\midrule
$\mathcal{L}\downarrow$                & 3.3438   & 1.8594 &
\cellcolor{green!15}{\textcolor{green!60!black}{\textbf{$L\downarrow~1.4844$}}} &
\cellcolor{green!15}{\textcolor{green!60!black}{\textbf{-44.4\%}}} \\

$L_{\text{recon}}\downarrow$           & 0.0412   & 0.0256 &
\cellcolor{green!15}{\textcolor{green!60!black}{\textbf{$L\downarrow~0.0156$}}} &
\cellcolor{green!15}{\textcolor{green!60!black}{\textbf{-37.9\%}}} \\

$L_{\text{sparse}}\downarrow$          & 0.2682   & 0.0209 &
\cellcolor{green!15}{\textcolor{green!60!black}{\textbf{$L\downarrow~0.2473$}}} &
\cellcolor{green!15}{\textcolor{green!60!black}{\textbf{-92.2\%}}} \\

$L_{\text{align}}\downarrow$           & 0.0605   & 0.0762 &
\cellcolor{red!15}{\textcolor{red!70!black}{\textbf{$L\uparrow~0.0157$}}} &
\cellcolor{red!15}{\textcolor{red!70!black}{\textbf{+26.0\%}}} \\

$L_{\text{ortho}}\downarrow$           & 0.5330   & 0.3760 &
\cellcolor{green!15}{\textcolor{green!60!black}{\textbf{$L\downarrow~0.1570$}}} &
\cellcolor{green!15}{\textcolor{green!60!black}{\textbf{-29.5\%}}} \\

$L_{\text{value}}\downarrow$           & 3.3169   & 3.4125 &
\cellcolor{red!15}{\textcolor{red!70!black}{\textbf{$L\uparrow~0.0956$}}} &
\cellcolor{red!15}{\textcolor{red!70!black}{\textbf{+2.9\%}}} \\

Diagonal Acc (train)$\uparrow$         & 0.9700   & 1.0000 &
\cellcolor{green!15}{\textcolor{green!60!black}{\textbf{$A\uparrow~0.0300$}}} &
\cellcolor{green!15}{\textcolor{green!60!black}{\textbf{+3.1\%}}} \\

Diagonal Acc (held-out)$\uparrow$      & 0.9699   & 0.9538 &
\cellcolor{red!15}{\textcolor{red!70!black}{\textbf{$A\downarrow~0.0161$}}} &
\cellcolor{red!15}{\textcolor{red!70!black}{\textbf{-1.7\%}}} \\

$L_{\text{indep}}\downarrow$           & 157.8025 & 4.7621 &
\cellcolor{green!15}{\textcolor{green!60!black}{\textbf{$L\downarrow~153.0404$}}} &
\cellcolor{green!15}{\textcolor{green!60!black}{\textbf{-97.0\%}}} \\
\bottomrule
\end{tabular}
\caption{Stage-1 ablation. Deltas are computed as \textbf{With--No}; losses prefer $\downarrow$ and accuracies prefer $\uparrow$.}
\label{tab:stage1_ablation}
\end{table}
\paragraph{Results.}
As shown in \cref{tab:stage1_ablation}, removing Stage~1 substantially worsens optimization and representation quality: total loss increases by $79.8\%$ ($1.86\!\rightarrow\!3.34$), with higher reconstruction error ($L_{\text{recon}}$: $+61.1\%$) and a large degradation in sparsity ($L_{\text{sparse}}$: $+1185.2\%$). Orthogonality also worsens ($L_{\text{ortho}}$: $+41.8\%$). In contrast, diagonal accuracy changes little ($+1.7\%$). Overall, Stage~1 is crucial for a stable sparse reconstructive basis, even if concept binding remains feasible without it.

\section{Results: Multi-Step Reasoning (2-hop)}
\label{sec:2hop}

Building on the layer sweep in \cref{fig:swap-heatmap}, we use Layer~6 for all 2-hop experiments, as it provides the most reliable substrate for slot-level binding and causal control in the step-wise setting (\cref{subsec:task_2hop}).

\subsection{Step-Wise Binding and Swap Controllability}

\Needspace*{16\baselineskip}
\begin{wrapfigure}[17]{r}{0.40\linewidth}
    \includegraphics[width=\linewidth]{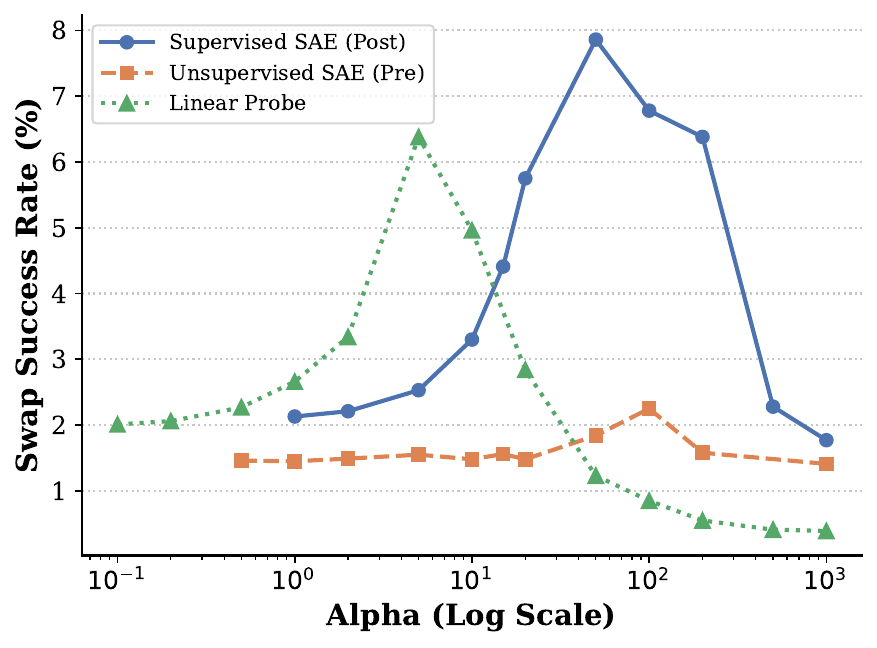}
    \caption{{Swap controllability in 2-hop reasoning.}
    The post-trained \modelname{} achieves $4\times$ higher swap success than the traditional SAE when  $\alpha\approx50$.}
    \label{fig:2hop_swap}
\end{wrapfigure}
\modelname{} achieves 100\% step-wise diagonal accuracy in 2-hop: the concept--slot confusion matrix is perfectly diagonal at both the $e_2$ step (binding to $r_1$) and the $e_3$ step (binding to $r_2$), confirming clean compositional alignment.

We next evaluate if these aligned slots are causal control knobs. Following the intervention protocol in \S\ref{subsec:swap_test}, we inject a decoded direction corresponding to a target relation slot with strength $\alpha$ during inference and measure whether the model's output switches to the entity implied by the swapped relation. \revision{We compare against a traditional SAE and a supervised linear probe~\citep{rimsky-etal-2024-steering}.
We train a linear probe to predict the relation label from the frozen hidden state using cross-entropy loss ($\sim100\%$ concept-classification accuracy). For swapping, we treat each classifier row weight $w_r$ as a concept direction and intervene directly in hidden space via
$h'_{\ell,t}=h_{\ell,t}-\alpha w_{r_{\text{input}}}+\alpha w_{r_{\text{target}}}$.}
As shown in \cref{fig:2hop_swap}, the supervised \modelname{} enables higher swap success than both the traditional SAE and linear probe, peaking near $\alpha\!\approx\!50$.
Detailed comparison against a standard SAE baseline and a fuller analysis of the 2-hop controllability gap are provided in \cref{app:twohop_gap}.

\begin{figure*}[t]
    \centering
    \includegraphics[width=\textwidth]{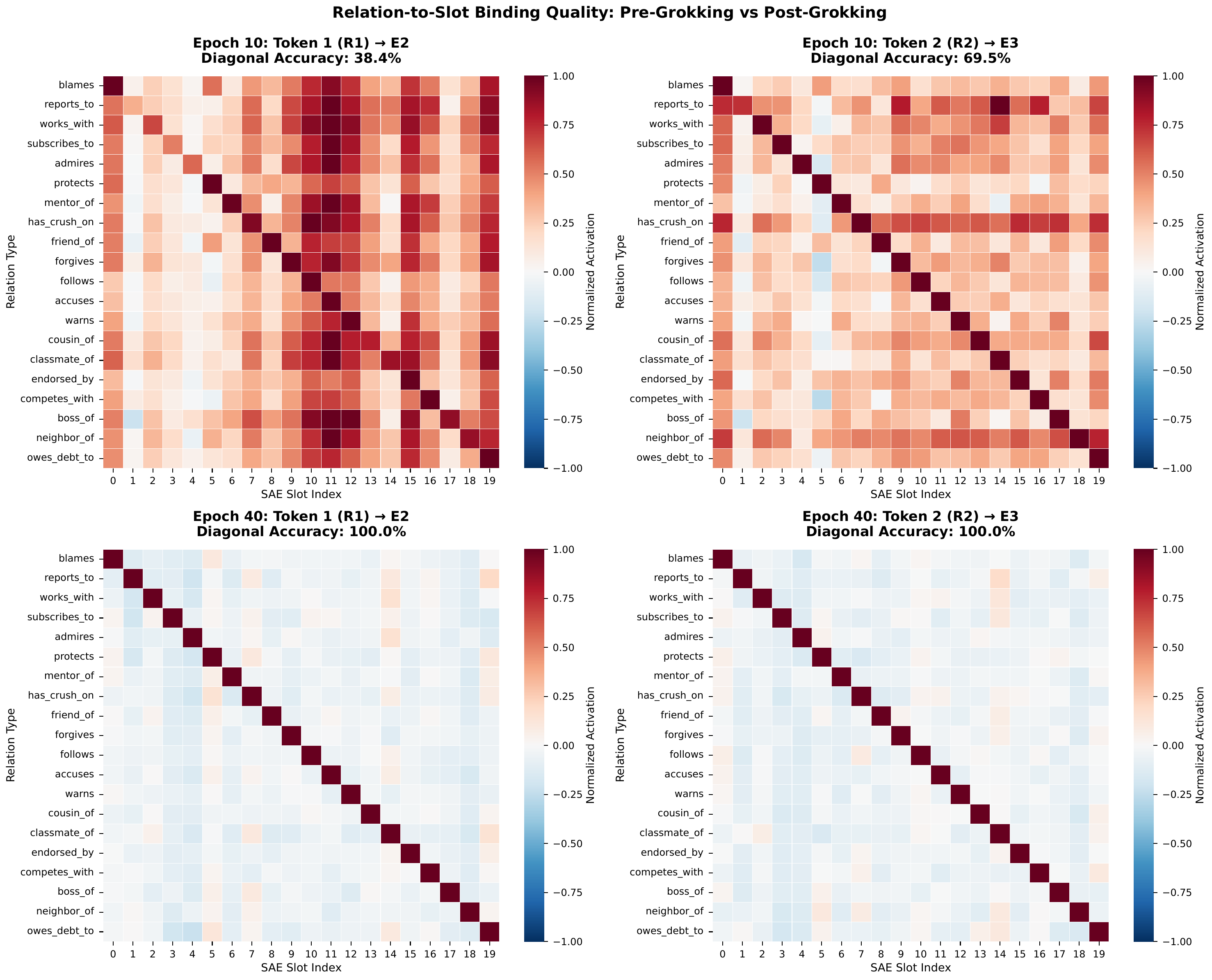}
    \caption{\textbf{Relation-to-slot binding (pre- vs.\ post-grokking).}
At steps $e_2$ and $e_3$, binding is initially entangled but becomes perfectly diagonal after grokking, indicating clean step-wise alignment.}
    \label{fig:2hop_binding_grokking_full}
\end{figure*}

\subsection{Grokking and Concept Binding}

We use \modelname{} as a diagnostic probe to understand grokking in 2-hop reasoning~\citep{wang2024grokked, ye2025how}. The base two-hop model exhibits grokking-like training dynamics: loss decreases smoothly, while validation accuracy remains low for many epochs before a sharp transition (\cref{fig:twohop_training_curves_grokking} in \cref{app:grokking}).

Our hypothesis is that \emph{before grokking}, concept evidence is present but not cleanly separated: the intended concept is encoded as a partially entangled mixture over slots. After the grokking transition, representations re-organize into structured, compositional features with stable step-wise 1-to-1 binding.

\begin{wrapfigure}{r}{0.42\linewidth}
    \centering
    \includegraphics[width=\linewidth]{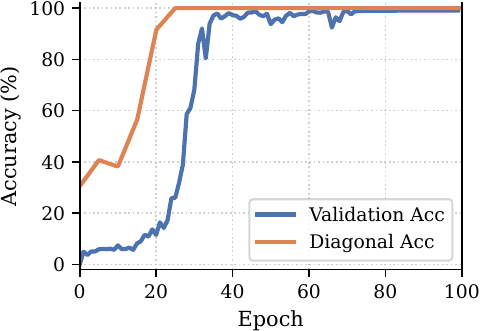}
    \caption{\textbf{Grokking-like emergence.}
    Diagonal accuracy saturates early, while validation accuracy improves later, suggesting delayed generalization in 2-hop reasoning.}
\label{fig:2hop_grokking_curve}
\vspace{-20pt}
\end{wrapfigure}

\looseness=-1 This predicts a \emph{lag between knowing and showing}. In \cref{fig:2hop_grokking_curve}, diagonal accuracy climbs rapidly and saturates early, whereas validation accuracy lags behind with a delayed jump. This dissociation suggests ``hidden'' progress: the model first organizes concept structure into stable internal features (``knowing''), and only subsequently learns to consistently \emph{use} that structure for out-of-distribution generalization (``showing'').

\Needspace*{17\baselineskip}
\cref{fig:2hop_binding_grokking_full} visualizes this transition at Token~1 (the $e_2$ step). Pre-grokking, binding remains diffuse (epoch~10: 38.4\%); post-grokking, the confusion matrix becomes perfectly diagonal (epoch~40: 100\%). Together with the delayed accuracy jump, these results are consistent with grokking as a phase transition from entangled representations to a clean, compositional regime where step-specific concepts become slot-addressable features that support multi-step generalization.

\noindent\textbf{Concurrent work.}
Recent concurrent work also uses SAEs to study grokking/emergent transitions: \citet{bereska2025superpositionlossycompressionmeasure} report sharp feature consolidation during grokking via an SAE-based superposition metric, while \citet{kumar2025predicting} use SAE co-activation graphs and find no evidence that global topology forecasts emergent jumps.
In contrast, we use \emph{concept-aligned} slots as a \emph{step-wise} diagnostic for how multi-hop evidence becomes compositional.

\section{Conclusion and Future Work}

In this work, we present \modelname{}, a framework that upgrades Sparse Autoencoders (SAEs) from descriptive probes to controllable interfaces.
To mitigate feature entanglement in unsupervised SAEs, we introduce a “pre-train, then post-train” curriculum that enforces explicit alignment between designated latents and an external ontology, while retaining a large residual bank for reconstruction.
This design yields reliable causal control, validated via paraphrase, robust concept swaps, and transfers cleanly to multi-hop reasoning.
\modelname{} also enables fine-grained analysis of representation learning dynamics.
We find a mechanistic account of grokking in compositional tasks: as generalization emerges, diffuse evidence concentrates into stable, step-wise bindings between relations and aligned slots.
Furthermore, by grounding abstract concepts in latent structure, \modelname{} moves SAE-based interpretability toward controllable internal representations.
\revision{A preliminary scalability study on FB15K (91 relations, \cref{app:fb15k}) confirms that concept alignment remains effective beyond the original synthetic settings.}
Detailed applications and future directions are provided in Appendix~\ref{app:future}.




    \section*{Limitations}
\label{sec:limitations}

\modelname{} is evaluated in a deliberately controlled setting on a frozen GPT-2, using synthetic 1-hop
biography QA and a 2-hop reasoning task rendered into natural-language prompts
(\cref{sec:experimental_setup}, \cref{sec:2hop}). This setup makes concept binding and causal
interventions precisely measurable (\cref{subsec:swap_result}) and allows us to extend
grokking-style mechanistic analysis to the \emph{concept-binding} level by tracking binding quality
across training (\cref{fig:2hop_grokking_curve} and \cref{fig:2hop_binding_grokking_full}). Unlike prior
work on grokked transformers \citep{wang2024grokked, ye2025how}, which train on explicit symbolic IDs, we intentionally avoid entity/relation identifiers and instead use natural-language templates to better mimic realistic usage and reduce shortcut learning; this
choice improves validity but also makes generalization harder and leaves open how our findings translate to ID-based settings. More broadly, our conclusions are currently limited to a GPT-2 backbone and templated data with paraphrase splits
(\cref{app:llm_templates}); highly indirect LLM-generated question styles can exceed the
backbone's QA capacity (\textsc{Unseen-Template} generalization; \cref{app:llm_templates}), so we focus on unseen but {direct}
template-level questions to ensure failures reflect representation-level alignment rather than
backbone incapability.
We also use a small ontology (\textit{e.g.,} 20 relations in 2-hop), which does not reflect full-scale SAE deployments; scaling to larger ontologies/models is left to future work, though the proposed binding and causal evaluation are not tied to this size.

Additionally, we evaluate on a single model architecture
(GPT-2) with synthetic data. While the controlled setting
enables rigorous evaluation of binding, disentanglement,
and controllability with unambiguous ground truth, it does
not capture the full complexity of natural language.
\modelname{}'s design around a predefined ontology of discrete
concepts is well-suited to domains with structured knowledge
bases (\textit{e.g.}, biomedicine, finance, law), where concepts are
deterministic and well-defined. Extending to fuzzier or
open-ended concept spaces---and to pre-trained models on
natural corpora---is an important direction for future work.
    \section*{Broader Impact}

Concept-aligned steering could in principle be misused to
induce targeted biases or factual errors. However, the same
capability is essential for \emph{detecting and correcting}
such biases. The transparency of named, inspectable concept
slots is itself a safety advantage over opaque, distributed
representations: interventions become auditable. Practical
mitigations include: (i)~\modelname{} requires model-internal
access and labeled ontology data, limiting casual misuse;
(ii)~the slot interface can be monitored for unauthorized
modifications; (iii)~the framework can be used
defensively---e.g., monitoring whether safety-relevant
concept slots are being suppressed during inference.
    \section*{Acknowledgments}

We thank the anonymous TMLR reviewers and the action editor for their valuable feedback. We gratefully acknowledge support from the University of Arizona Undergraduate Research Travel Grant, which provided funding for Minglai Yang. We also thank the College of Information Science for additional student research funding and the AI Club at University of Arizona for their support.

\bibliography{main}
\bibliographystyle{tmlr}

    \clearpage

\appendix
\section*{Content of Appendix}
\begin{itemize}[noitemsep, topsep=0pt]
    \item[\ref{app:dataset}] Biography Dataset Generation
    \item[\ref{app:training}] Base Language Model Training
    \item[\ref{app:sae}] Supervised Sparse Autoencoder
    \item[\ref{app:metrics}] Evaluation Metrics
    \item[\ref{app:sae_visualizations}] SAE Feature Activations by Relation Type
    \item[\ref{app:qual_swaps}] Additional Qualitative Swap Examples
    \item[\ref{app:mig}] Mutual Information Gap Analysis
    \item[\ref{app:latent_vis}] Latent Space Visualization (t-SNE)
    \item[\ref{app:side_effects}] Steering Side-Effects Analysis
    \item[\ref{app:fb15k}] Scalability: FB15K with 91 Relations
    \item[\ref{app:twohop_gap}] Standard SAE Baseline and 2-Hop Controllability Gap
    \item[\ref{app:future}] Applications and Future Directions
    \item[\ref{app:twohop_format}] Two Hop Reasoning Dataset Construction
    \item[\ref{app:grokking}] Two-Hop Training Dynamics
\end{itemize}

\section{Biography Dataset Generation}
\label{app:dataset}

\subsection{Synthetic Biography Dataset}

We generated 1,000 synthetic person profiles, each containing six factual attributes: birth date, birth city, university, major, employer, and work city. Each person was paired with 5 biography variants constructed from template-based generation. For question-answering evaluation, we employed a template-split strategy: templates 0--1 were designated for training (in-distribution), while templates 2--3 served as out-of-distribution test cases to evaluate semantic generalization beyond pattern matching.

\subsection{Entity Vocabulary}

The dataset drew from the following entity sets:

\begin{itemize}[leftmargin=*,noitemsep]
    \item \textbf{First names:} 411 diverse given names spanning traditional and modern choices
    \item \textbf{Middle names:} 461 names used for middle name generation
    \item \textbf{Last names:} 1,002 surnames representing common American family names
    \item \textbf{Birth cities:} 50 major U.S. cities (\emph{e.g.}, New York, Los Angeles, Chicago, etc.)
    \item \textbf{Universities:} 341 U.S. colleges and universities spanning liberal arts colleges, research universities, technical institutes, and military academies
    \item \textbf{Majors:} 101 academic fields ranging from STEM disciplines (Computer Science, Mechanical Engineering, Biochemistry) to humanities (Philosophy, Art History, Creative Writing) and professional programs (Business Administration, Nursing, Architecture)
    \item \textbf{Companies:} 327 major U.S. corporations with associated headquarters cities, covering diverse industries including technology, finance, healthcare, retail, and manufacturing
\end{itemize}

This vocabulary size enables the generation of approximately $411 \times 461 \times 1{,}002 \times 50 \times 341 \times 101 \times 327 \approx 1.04 \times 10^{17}$ unique person profiles, ensuring minimal memorization pressure and focusing evaluation on semantic understanding rather than rote learning. We pick 1000 samples (6000 entities) in our dataset for the factual knowledge memorization.

\subsection{Question-Answer Templates}
\label{app:qa_pairs}
Each of the six semantic relations was probed using four distinct question templates, enabling controlled evaluation of template generalization. Table~\ref{tab:qa_templates} lists all templates used in our experiments.

\begin{table*}[t]
\centering
\small
\setlength{\tabcolsep}{6pt}
\begin{tabular}{@{}l>{\raggedright\arraybackslash}p{6.2cm}>{\raggedright\arraybackslash}p{6.2cm}@{}}
\toprule
\textbf{Relation} & \textbf{Train templates (T0--T1)} & \textbf{Held-out templates (T2--T3)} \\
\midrule
Birth Date
& When was \{FULL\_NAME\} born? \newline
On what date was \{FULL\_NAME\} born?
& What is \{FULL\_NAME\}'s birth date? \newline
Can you tell me the birth date of \{FULL\_NAME\}? \\
\midrule
Birth City
& Where was \{FULL\_NAME\} born? \newline
In what city was \{FULL\_NAME\} born?
& What is \{FULL\_NAME\}'s birth city? \newline
Can you tell me the birth city of \{FULL\_NAME\}? \\
\midrule
University
& Where did \{FULL\_NAME\} go to college? \newline
Which college did \{FULL\_NAME\} attend?
& What is \{FULL\_NAME\}'s alma mater? \newline
Which university did \{FULL\_NAME\} attend? \\
\midrule
Major
& What was \{FULL\_NAME\}'s major? \newline
What is \{FULL\_NAME\}'s field of study?
& What did \{FULL\_NAME\} study? \newline
What field did \{FULL\_NAME\} study in? \\
\midrule
Employer
& Who does \{FULL\_NAME\} work for? \newline
What company does \{FULL\_NAME\} work for?
& What is \{FULL\_NAME\}'s employer? \newline
Which company employs \{FULL\_NAME\}? \\
\midrule
Work City
& Where does \{FULL\_NAME\} work? \newline
What city does \{FULL\_NAME\} work in?
& What is \{FULL\_NAME\}'s work city? \newline
In which city is \{FULL\_NAME\} employed? \\
\bottomrule
\end{tabular}
\caption{Template split for surface-form robustness. T0--T1 are used for training, while T2--T3 are held out for \textsc{Unseen-Template} evaluation; the split is designed to reduce lexical overlap in relation triggers (\textit{e.g.}, \emph{born} vs.\ \emph{birth date}, \emph{work for} vs.\ \emph{employer}).}
\label{tab:qa_templates}
\end{table*}

This template design ensures semantic diversity while maintaining consistent information content, allowing us to test whether the SAE captures abstract semantic relations rather than surface-level linguistic patterns.

\subsection{Using an LLM to generate unseen questions}
\label{app:llm_templates}

Beyond the fixed template split (T0--T3; \cref{tab:qa_templates}), we explored generating additional
\emph{unseen} questions with an external LLM. Concretely, for each 1-hop RC query $(e_1,r)$ (entity $e_1$
and relation $r\in\mathcal{R}$), we few-shot prompt Claude~3.5~Sonnet via the OpenRouter API with two instantiated in-distribution questions (from T0--T1 for the same relation) and ask it to produce exactly two new, \emph{simple and direct} paraphrases that query the same attribute. We enforce basic constraints (use the
exact entity name, keep questions short, return exactly two lines); the full prompt is shown in
\cref{fig:ood_prompt}. This procedure produces \emph{instance-specific} unseen questions (unique per $(e_1,r)$), rather than global templates shared across entities.

\begin{figure*}[!htbp]
\centering
\fbox{\parbox{0.97\textwidth}{\footnotesize
\textbf{User prompt used for OOD question generation (verbatim, with placeholders).}\\[4pt]
{\ttfamily\raggedright
You are helping generate simple question variations for a dataset. Your task is to create 2 NEW questions that ask about \{RELATION\_DESC\} for the person "\{FULL\_NAME\}".\\[4pt]
IMPORTANT REQUIREMENTS:\\
1. Keep questions SIMPLE and DIRECT - similar to these training examples but with SLIGHT variations:\\
\ \ \ - "\{ID\_EXAMPLE\_1\}"\\
\ \ \ - "\{ID\_EXAMPLE\_2\}"\\[4pt]
2. Make SMALL changes only - change just 1-2 words or reorder slightly\\
3. Questions must be answerable with a direct factual answer (date, city, university name, etc.)\\
4. DO NOT ask questions that require explanation or opinion\\
5. Use the exact name "\{FULL\_NAME\}" in each question\\
6. Keep questions SHORT (under 15 words)\\[4pt]
Generate exactly 2 simple question variations, one per line. Do NOT number them or add any other text.
}}
}
\caption{Few-shot prompt used to generate OOD paraphrases for each $(p,r)$ pair. \{RELATION\_DESC\} is a short natural-language description of the target relation (\textit{e.g., ``the person's birth date''}), and \{ID\_EXAMPLE\_1\}--\{ID\_EXAMPLE\_2\} are instantiated from the ID training templates (T0--T1).}
\label{fig:ood_prompt}
\end{figure*}

In practice, many generations drift toward more indirect or complex phrasings that the base GPT-2 backbone cannot reliably answer, even when the questions are semantically well-formed. To isolate the effect of concept alignment (rather than backbone capacity), we therefore restrict the main experiments to the controlled template split in \cref{tab:qa_templates} and treat LLM-generated unseen questions as an exploratory stress test.

\paragraph{Examples.}
Compared to the fixed templates, Claude~3.5~Sonnet often introduces indirect phrasing. For \textsc{birth\_date}:
(i) \textit{``Do you happen to know the calendar date that marks Todd Raul Hanson's arrival?''} and
(ii) \textit{``I'm trying to find out the calendar date on which Edith Rocky Taylor made her first appearance.''}
(answer: \texttt{12,August,1991}). For \textsc{employer}:
\textit{``I'm curious about the business where Jonathan Kiera Carney earns their living -- do you know it?''}
(answer: \texttt{Caterpillar Inc.}).

\FloatBarrier
\section{Base Language Model Training}
\label{app:training}

\subsection{Model Architecture}

We employed GPT-2 with 124M parameters as the base causal language model, featuring 768-dimensional hidden representations and 12 transformer layers.

\subsection{Training Objective}

The model was trained using a two-component curriculum: (1) \textit{biography memorization}, where the model learned to predict entire biography sequences, and (2) \textit{pure question-answering}, where only answer tokens contributed to the loss while question prompts were masked (label = -100).

\subsection{Optimization Hyperparameters}

\paragraph{Selection protocol.}
We did {not} perform an extensive hyperparameter search for base LM training. Unless otherwise
noted, we {mainly} adopt the optimization settings reported in \cite{AL2023-knowledge1}
(AdamW with warmup, cosine decay, and gradient clipping) and keep them fixed across all runs. We only
ran minimal sanity checks to ensure training stability (no divergence) and to achieve near-saturated
train/validation performance on our synthetic curriculum; we did not tune hyperparameters to optimize
downstream controllability or binding metrics.

Training was conducted with the following configuration:

\begin{itemize}[leftmargin=*,noitemsep]
    \item \textbf{Maximum training steps:} 80,000
    \item \textbf{Effective batch size:} 96 (distributed across available GPUs)
    \item \textbf{Learning rate schedule:} Linear warmup to $1 \times 10^{-3}$ over 1,000 steps, followed by cosine annealing to $1 \times 10^{-4}$
    \item \textbf{Optimizer:} AdamW with weight decay 0.1 and $\epsilon = 1 \times 10^{-6}$
    \item \textbf{Gradient clipping:} Maximum norm 1.0
    \item \textbf{Maximum sequence length:} 512 tokens
    \item \textbf{Checkpoint frequency:} Every 10,000 steps
\end{itemize}

\subsection{Activation Collection}
\label{app:activation_collection}

Hidden states were extracted from the residual stream at the final token position of the question prompt (immediately before answer generation), representing the point where the model ``decides'' what information to retrieve. We collected activations from all 12 transformer layers independently to analyze the emergence of semantic binding across network depth.

\section{Concept-Aligned Sparse Autoencoder}
\label{app:sae}

\subsection{Model Architecture}

Our supervised SAE extends a standard sparse autoencoder with $|\mathcal{R}|$ dedicated relation slots, while delegating residual variance to a large bank of free features. The configuration is:

\begin{itemize}[leftmargin=*,noitemsep]
    \item \textbf{Input dimension:} $d{=}768$ (GPT-2 hidden size)
    \item \textbf{Total latent features:} $K{+}|\mathcal{R}|$ with $K\in\{10000, 100000\}$ (task-dependent) and $|\mathcal{R}|$
    \begin{itemize}[noitemsep]
        \item \textit{Free features:} $z_{\text{rest}}\in\mathbb{R}^{K}$ (unsupervised)
        \item \textit{Relation slots:} $z_{\text{concept}}\in\mathbb{R}^{|\mathcal{R}|}$ (supervised; one per relation)
    \end{itemize}
    \item \textbf{Encoder:} $E:\mathbb{R}^{d}\!\to\!\mathbb{R}^{|\mathcal{R}|+K}$, \;
    $z=\mathrm{ReLU}(W_e h+b_e)$ with $z=[z_{\text{concept}};z_{\text{rest}}]$
    \item \textbf{Decoder:} $D:\mathbb{R}^{|\mathcal{R}|+K}\!\to\!\mathbb{R}^{d}$,\;
    $\hat h=W_d z+b_d$
\end{itemize}

\paragraph{Binding Mechanism.}
We partition $z=[z_{\text{concept}};z_{\text{rest}}]$ with $z_{\text{concept}}\in\mathbb{R}^{|\mathcal{R}|}$ and
define $p(r\mid x)=\mathrm{softmax}(z_{\text{concept}})$.
The binding loss applies cross-entropy with the one-hot gold relation, encouraging the correct relation slot to dominate.

\subsection{Multi-Stage Training Protocol}

Training proceeds in two stages to ensure stable convergence:

\begin{itemize}[leftmargin=*,noitemsep]
    \item \textbf{Stage 1 (Reconstruction-Only):} 100 epochs focusing solely on autoencoding quality before introducing binding constraints
    \item \textbf{Stage 2 (Full Supervision):} 500 epochs with complete loss function
\end{itemize}

\subsection{Loss Function}
\label{app:loss_function}
The total training objective (see Section~\ref{sec:implementation} of the main paper) combines six components. In Stage~1 (100 epochs), only reconstruction loss is active to stabilize the encoder-decoder. In Stage~2 (500 epochs), all six losses are jointly optimized:
\begin{align}
\mathcal{L}_{\text{total}} &= \lambda_{\text{recon}} \mathcal{L}_{\text{recon}} + \lambda_{\text{sparse}} \mathcal{L}_{\text{sparse}} + \lambda_{\text{align}} \mathcal{L}_{\text{align}} \nonumber \\
&\quad 
+ \lambda_{\text{ortho}} \mathcal{L}_{\text{ortho}} + \lambda_{\text{value}} \mathcal{L}_{\text{value}}
\end{align}

Each component serves a specific purpose:

\paragraph{Reconstruction Loss.} 
\begin{equation}
\mathcal{L}_{\text{recon}} = \text{MSE}(\hat{h}, h) = \frac{1}{d} \sum_{i=1}^{d} (\hat{h}_i - h_i)^2
\end{equation}
Measures mean squared error between original activation $h$ and reconstructed activation $\hat{h} = W_{\text{dec}} \cdot z$, where $d=768$ is the hidden dimension. This ensures the SAE preserves information necessary for the language model's downstream predictions while learning a compressed latent representation.

\paragraph{Sparsity Loss.} 
\begin{equation}
\mathcal{L}_{\text{sparse}} = \frac{1}{B \cdot n_{\text{free}}} \sum_{b=1}^{B} \sum_{j=1}^{n_{\text{free}}} |z_{b,j}|
\end{equation}
Enforces L1 penalty on the large number of free slots across batch size $B$, encouraging the model to activate only a small subset of features per sample. Sparse activations improve interpretability by ensuring each latent feature captures distinct semantic properties rather than distributing information diffusely.

\paragraph{Alignment Loss.} 
\begin{align}
\mathcal{L}_{\text{align}}
&= \text{CrossEntropy}(\text{softmax}(z_{\text{rel}}), y) \\
&= -\frac{1}{B} \sum_{b=1}^{B} \sum_{r=1}^{6}
    y_{b,r}
    \log 
    \frac{
        \exp\!\big(z_{b, n_{\text{free}}+r}\big)
    }{
        \sum_{r'=1}^{6} \exp\!\big(z_{b, n_{\text{free}}+r'}\big)
    }.
\end{align}
Provides supervised guidance where $y$ is a one-hot vector with $y_{b,\text{rule\_idx}_b} = 1$ indicating the ground-truth relation type. This cross-entropy loss over softmax-normalized relation slot activations enforces that the 6 relation slots (indices $n_{\text{free}}+1$ through $n_{\text{free}}+6$) form a probability distribution with mass concentrated on the correct semantic relation. This is the binding loss used in all main experiments, enabling explicit classification of question types to specific latent dimensions.

\paragraph{Orthogonality Loss.} 
\begin{equation}
\mathcal{L}_{\text{ortho}} = \sum_{r=1}^{6} \sum_{j=1}^{n_{\text{free}}} \left( \frac{1}{B} \sum_{b=1}^{B} (z_{b,n_{\text{free}}+r} - \bar{z}_r)(z_{b,j} - \bar{z}_j) \right)^2
\end{equation}
Enforces statistical independence between supervised relation slots and unsupervised free slots by minimizing their cross-covariance. This prevents relation slots from encoding information already captured by free features, ensuring clean separation between task-specific and general-purpose representations.

\paragraph{Value Prediction Loss.}
To encourage each relation slot to be {task-informative}, \textit{i.e.}, to carry evidence that supports answer
prediction beyond merely signaling its identity, we add an auxiliary value objective. For each training
example $b$, we take the activation of the gold relation slot $r_b=\text{rule\_idx}_b$ and predict the first
token of the answer via a relation-specific head:
\begin{equation}
\mathcal{L}_{\text{value}}
= \frac{1}{B} \sum_{b=1}^{B}
\text{CrossEntropy}\Big(
  \text{V}_{r_b}\!\big(z_{b,n_{\text{free}}+r_b}\big),
  t_b
\Big),
\end{equation}
where $t_b$ is the first token of the ground-truth answer and $\text{V}_{r}$ is a two-layer MLP mapping the scalar slot activation to vocabulary logits. We backpropagate
$\mathcal{L}_{\text{value}}$ jointly with the reconstruction and binding losses so that concept slots remain
predictive of the target output, while the large free-feature bank captures residual variation needed for
high-fidelity reconstruction. The value heads serve as an auxiliary training signal (and diagnostic readout)
and are not used in inference-time interventions.

\paragraph{Independence validation (not trained).}
Although we do not optimize an explicit free-slot independence term, we validate redundancy via an \emph{independence score} computed as the squared off-diagonal covariance among free features:
\begin{equation}
\mathcal{L}_{\text{indep}} = \sum_{i \neq j} \left( \frac{1}{B} \sum_{b=1}^{B} (z_{b,i} - \bar{z}_i)(z_{b,j} - \bar{z}_j) \right)^2,
\end{equation}
where $\bar{z}_i = \frac{1}{B}\sum_{b=1}^{B} z_{b,i}$.
Without Stage~1, this score explodes ($4.76\!\rightarrow\!157.80$, $+3213.7\%$), indicating highly correlated and redundant free features.

\subsection{Loss Weight Selection}

\paragraph{Selection protocol.}
We did {not} perform an extensive hyperparameter search. Unless explicitly noted, we use
\textbf{standard/default SAE choices} and keep them fixed across all runs. We only ran minimal
sanity checks to avoid degenerate behavior (\textit{e.g.}, dead features or collapse), and we did not tune
weights to maximize reported metrics.

The loss components are weighted to balance competing objectives:

\begin{itemize}[leftmargin=*,noitemsep]
    \item $\lambda_{\text{recon}} = 1.0$ --- High priority is given to faithful reconstruction to maintain model performance. This ensures the SAE does not distort the information flow through the network.
    
    \item $\lambda_{\text{sparse}} = 1 \times 10^{-3}$ --- Gentle L1 penalty on free slots. This small weight prevents over-suppression of activations while still encouraging selective feature usage. Stronger sparsity penalties ($\lambda > 10^{-2}$) caused excessive dead neurons and degraded reconstruction quality in preliminary experiments.
    
    \item $\lambda_{\text{align}} = 1.0$ --- Strong supervision signal to ensure reliable slot-relation binding. This weight is balanced with reconstruction to achieve $>95\%$ diagonal accuracy on in-distribution templates in \cref{fig:combined_binding_swap}.
    
    
    \item $\lambda_{\text{ortho}} = 1 \times 10^{-2}$ --- Moderate orthogonality constraint between relation and free slots. This maintains separation between supervised and unsupervised features, preventing information leakage that could compromise the interpretability of relation slots.
    
    \item $\lambda_{\text{value}} = 0.5$ --- Balanced weight for answer prediction. This auxiliary task provides a training signal to ensure relation slots encode semantically meaningful information, but is weighted lower than alignment to avoid dominating the optimization.
\end{itemize}

\subsection{Training Hyperparameters}

\paragraph{Selection protocol.}
We use standard/default optimizer settings (AdamW) and do not tune
hyperparameters for peak performance; we only ran minimal sanity checks for training stability.

\paragraph{Optimizer configuration.}
We use AdamW with the following settings~\citep{loshchilov2019decoupledweightdecayregularization}:
\begin{itemize}[leftmargin=*,noitemsep]
    \item \textbf{Learning rate:} $1 \times 10^{-3}$ (constant, no warmup or decay)
    \item \textbf{Weight decay:} $0.0$ (L2 regularization disabled to avoid interfering with explicit sparsity constraints)
    \item \textbf{Betas:} $(0.9, 0.999)$ (default momentum coefficients)
    \item \textbf{Epsilon:} $1 \times 10^{-8}$ (numerical stability constant)
    \item \textbf{Batch size:} 64 samples per update
    \item \textbf{Gradient clipping:} None (training was stable without clipping)
\end{itemize}

\paragraph{Training Schedule.} The constant learning rate without decay was chosen because the two-stage training protocol naturally provides curriculum learning: Stage~1 establishes a good initialization for the encoder-decoder using traditional SAE framework~\cite{shu-etal-2025-survey}, after which Stage~2 refines the latent structure, as defined as SAE post-training. Preliminary experiments with cosine annealing showed no improvement over constant learning rate for this setting.

\section{Evaluation Metrics}
\label{app:metrics}

This section provides detailed mathematical definitions for all evaluation metrics reported in Section~5 of the main paper.

\subsection{Concept--Slot Binding Metrics}

We evaluate the quality of concept--slot binding using the following metrics, reported in \cref{tab:layer_comparison} and \cref{fig:bind} of the main paper:

\paragraph{Diagonal Accuracy.} Our primary binding metric (\cref{subsec:metrics}). It quantifies the one-to-one mapping quality between ground-truth relations and predicted slots using the relation--slot confusion matrix:
\begin{equation}
\text{Diag} = \frac{1}{R} \sum_{i=1}^{R} C_{ii}
\end{equation}
where $R=|\mathcal{R}|$ is the number of relations ($R=6$ in the 1-hop setting) and $C$ is the normalized confusion matrix with $C_{ij} = \#\{r=i,\hat{r}=j\}/\#\{r=i\}$, where $\hat{r} = \operatorname*{argmax}_{j}\, z_{\text{rel},j}$ is the top-activated relation slot at the answer token and $z_{\text{rel}}^{(i)}$ is the relation slot activation vector for question $i$ ($i=1,\dots,N$). Perfect binding yields $\text{Diag} = 1.0$, while random assignment gives $\text{Diag} \approx 1/R$ ($\approx 0.167$ for $R=6$). Confusion matrices are visualized in \cref{fig:bind} for layer-wise analysis.

\paragraph{Margin.} The logit difference between the top-1 and top-2 slot predictions, measuring binding confidence:
\begin{equation}
\text{Margin} = \frac{1}{N} \sum_{i=1}^{N} (z_{\text{rel}, j_1}^{(i)} - z_{\text{rel}, j_2}^{(i)})
\end{equation}
where $j_1$ and $j_2$ are the indices of the highest and second-highest activations. Higher margins indicate more confident and unambiguous binding.

\paragraph{Answer Accuracy.} Exact-match accuracy for generated answers:
\begin{equation}
\text{Acc}_{\text{answer}} = \frac{1}{N} \sum_{i=1}^{N} \mathbb{1}[\hat{a}_i = a_i]
\end{equation}
where the normalization function handles multiple date formats (\textit{e.g.}, \textit{``Day, Month, Year''}; \textit{``Month Day, Year''}; \textit{``YYYY-MM-DD''}) to avoid penalizing formatting differences.

\paragraph{Swap Intervention Accuracy (Causal Control).} Measures the full language model's answer generation after latent manipulation in swap experiments (\cref{subsec:swap_result}). After modifying relation slot activations ($z_i^{\text{orig}} \leftarrow 0$, $z_j^{\text{target}} \leftarrow \alpha$), we decode to obtain $\hat{h} = W_{\text{dec}} \cdot z'$ and feed this modified activation through the remaining transformer layers to generate text using the LLM's standard autoregressive generation. The value heads are \textit{not used} in these experiments. This validates causal control over model behavior. Optimal swap success: 85\% at $\alpha \approx 2$ (Layer 6).

The swap intervention accuracy validates that these representations causally influence the full model's behavior.

\subsection{Reconstruction Quality}

We measure the fidelity of the autoencoder's reconstruction using mean squared error:
\begin{equation}
\text{MSE}_{\text{recon}} = \frac{1}{N} \sum_{i=1}^{N} \|h^{(i)} - \hat{h}^{(i)}\|^2
\end{equation}
where $h^{(i)}$ is the original 768-dimensional activation vector from GPT-2's residual stream and $\hat{h}^{(i)} = W_{\text{dec}} \cdot z^{(i)}$ is the reconstructed activation after encoding and decoding through the SAE. The reconstruction target is the raw activation, not normalized or preprocessed. Layer 6 achieves MSE $\approx 7.42 \times 10^{-2}$, representing a trade-off between reconstruction fidelity and semantic structure---early layers achieve lower MSE (\textit{e.g.}, Layer 0: $6.53 \times 10^{-5}$) but lack meaningful concept binding. The relatively higher MSE in middle layers reflects the cost of enforcing interpretable slot structure while maintaining sufficient information for downstream task performance.

\subsection{Swap Controllability}
\label{subsec:swap_controllability}

To test whether ontology-aligned slots serve as usable \emph{control knobs}, we measure
the success rate of answer swaps under inference-time steering along decoded SAE
directions (protocol in \cref{subsec:swap_test}; implementation details in~\cref{app:activation_collection}).

\noindent\textbf{Intervention.}
Let $h_{\ell,t}\in\mathbb{R}^d$ denote the residual stream at layer $\ell$ and token position $t$,
and let $W_{\text{dec}}\in\mathbb{R}^{d\times K}$ be the SAE decoder. For a target concept slot
$j\in\{1,\dots,K\}$, define the decoded direction $v_j := W_{\text{dec}} e_j$.
Given a question whose gold relation is $r^\star$ and a chosen target slot $j\neq \pi(r^\star)$,
we steer the residual stream by
\begin{equation}
h'_{\ell,t} \;=\; h_{\ell,t} + \alpha v_j
\;=\; h_{\ell,t} + W_{\text{dec}}(\alpha e_j),
\end{equation}
then continue the forward pass through the remaining transformer layers and decode the answer.
We sweep amplification strengths
\begin{equation}
\alpha \in \{0.1, 0.5, 1, 2, 5, 10, 20, 50, 100, 200, 500, 1000\}.
\end{equation}

\noindent\textbf{Swap success.}
For swap trial $m$, let $p_m$ be the subject entity and let $r^{\text{tgt}}_m$ be the target relation
corresponding to the amplified slot $j$. Let $\hat{a}^{\text{swap}}_m$ be the generated answer
under the intervention. We count the swap as successful if the model outputs the gold value for
the \emph{target} relation, $g(p_m, r^{\text{tgt}}_m)$:
\begin{equation}
\mathrm{Swap}_{\alpha}
= \frac{1}{M} \sum_{m=1}^{M}
\mathbb{1}\!\big[\hat{a}_m^{\text{swap}} = g(p_m, r^{\text{tgt}}_m)\big].
\end{equation}

As shown in Figure~\ref{fig:swap-heatmap}, controllability is maximized at moderate
amplification ($\alpha \approx 2$): at Layer~6, swap success rises from $0.54$ at $\alpha=1$
to $0.85$ at $\alpha=2$, but drops to $0.23$ at $\alpha=10$, indicating that over-amplification
destabilizes the intervention.
\section{Layer-wise SAE Feature Comparison}
\label{app:sae_visualizations}

\begin{figure*}[p]
\centering
\begin{subfigure}[t]{0.48\textwidth}
    \centering
    \includegraphics[width=\linewidth]{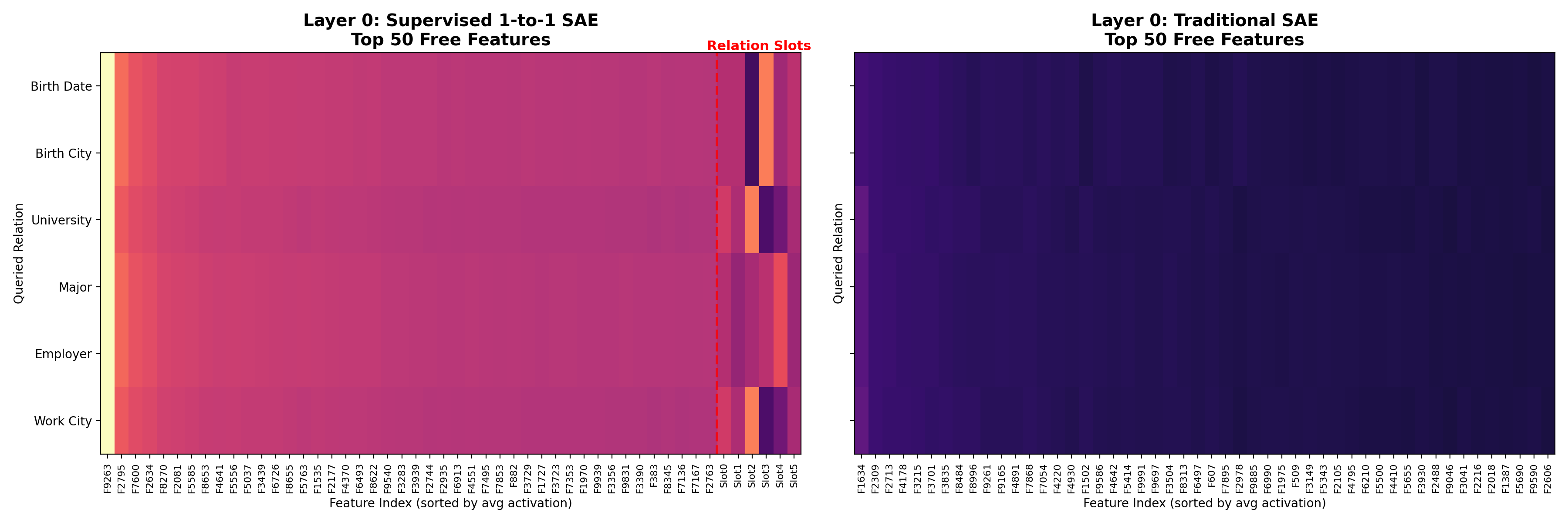}
    \caption{Layer 0}
    \label{fig:sae_layer00}
\end{subfigure}
\hfill
\begin{subfigure}[t]{0.48\textwidth}
    \centering
    \includegraphics[width=\linewidth]{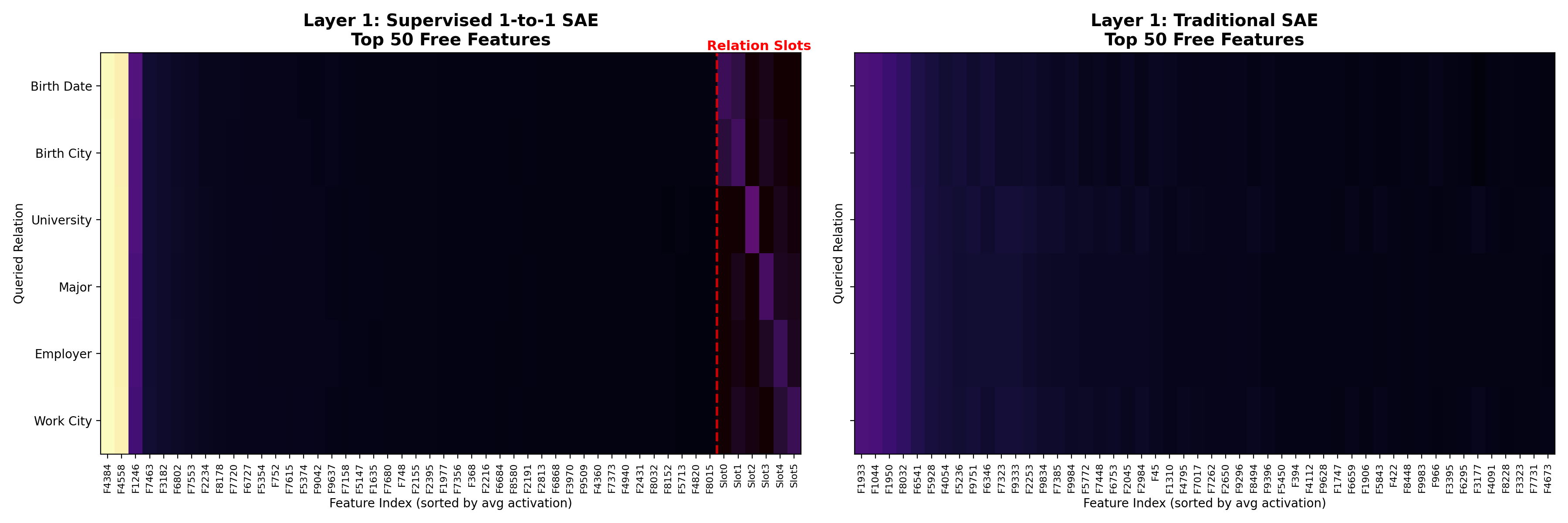}
    \caption{Layer 1}
    \label{fig:sae_layer01}
\end{subfigure}

\vspace{0.15cm}

\begin{subfigure}[t]{0.48\textwidth}
    \centering
    \includegraphics[width=\linewidth]{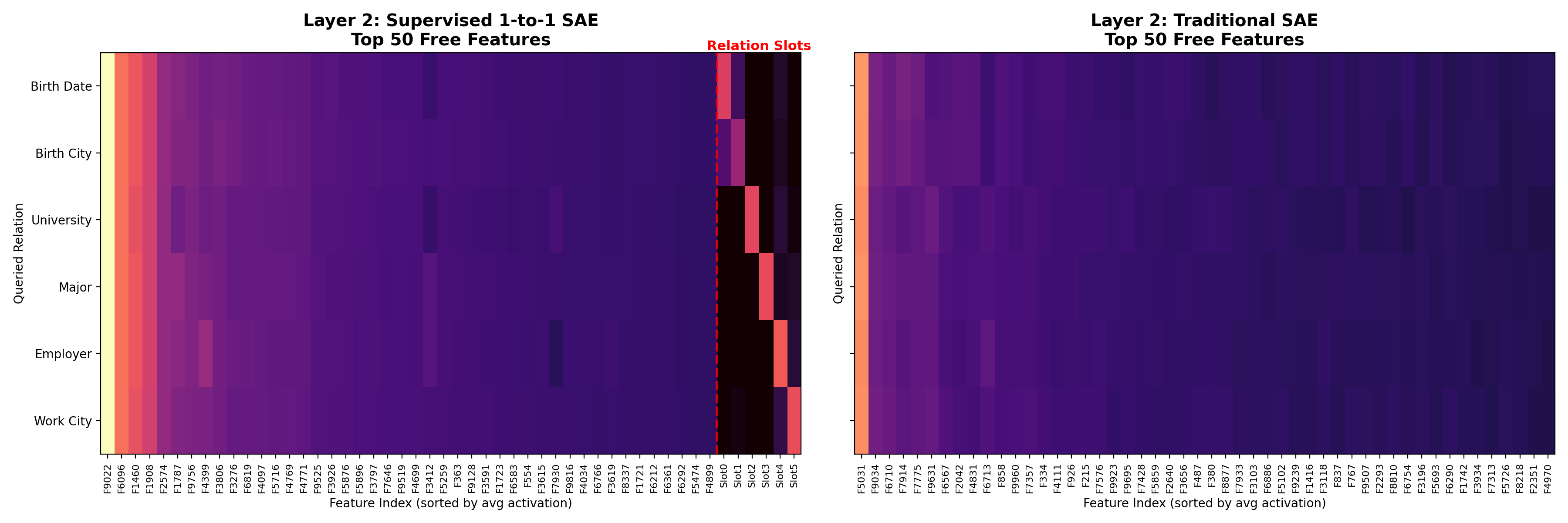}
    \caption{Layer 2}
    \label{fig:sae_layer02}
\end{subfigure}
\hfill
\begin{subfigure}[t]{0.48\textwidth}
    \centering
    \includegraphics[width=\linewidth]{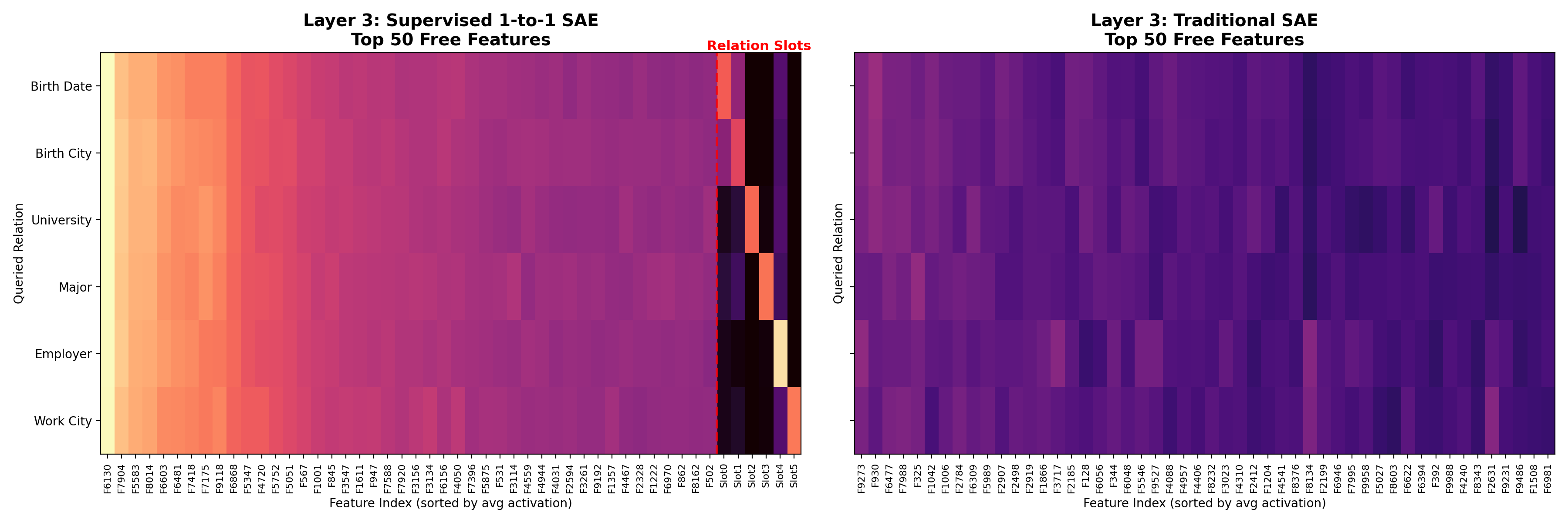}
    \caption{Layer 3}
    \label{fig:sae_layer03}
\end{subfigure}

\vspace{0.15cm}

\begin{subfigure}[t]{0.48\textwidth}
    \centering
    \includegraphics[width=\linewidth]{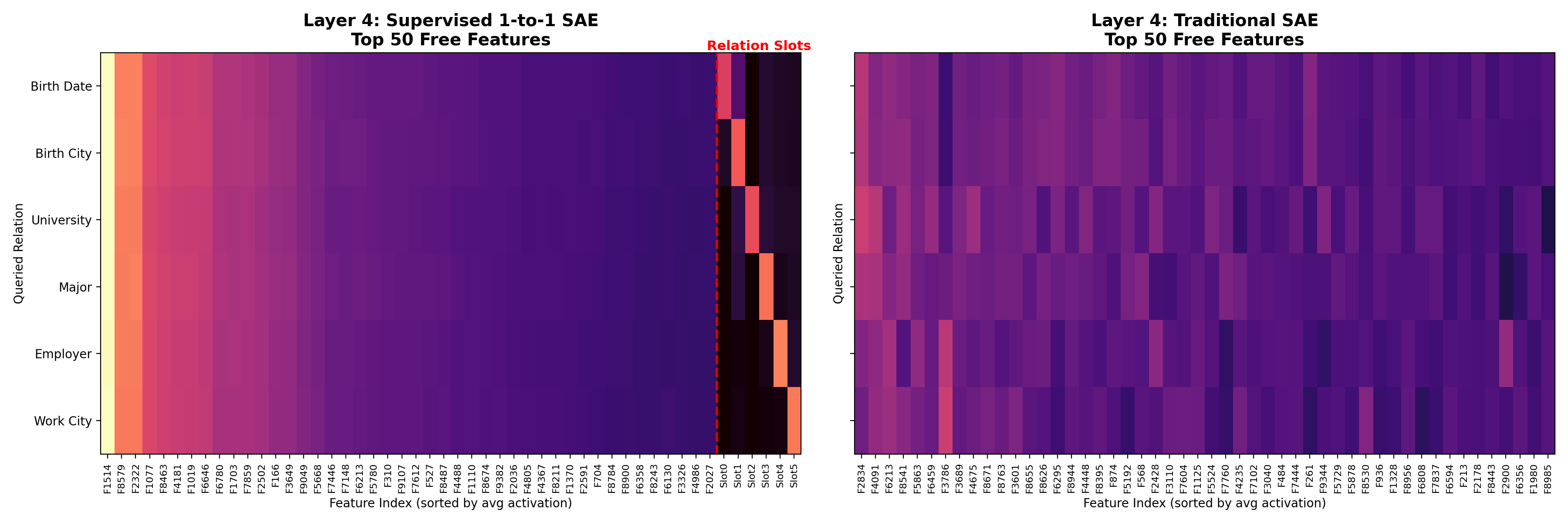}
    \caption{Layer 4}
    \label{fig:sae_layer04}
\end{subfigure}
\hfill
\begin{subfigure}[t]{0.48\textwidth}
    \centering
    \includegraphics[width=\linewidth]{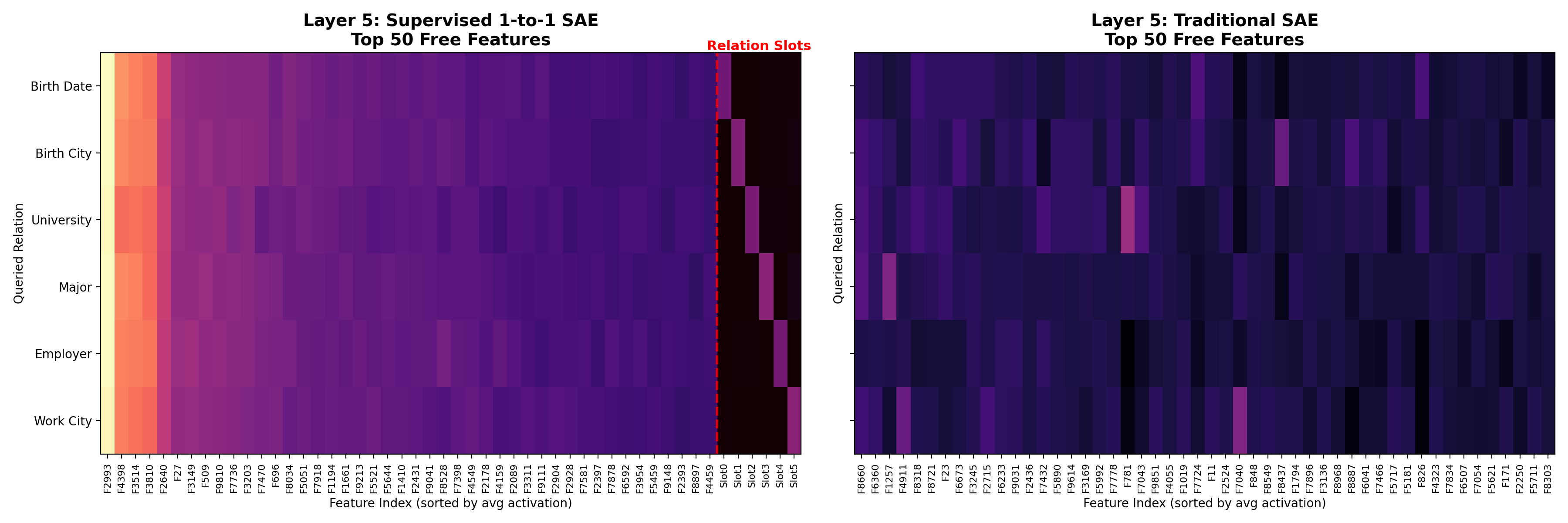}
    \caption{Layer 5}
    \label{fig:sae_layer05}
\end{subfigure}

\vspace{0.15cm}

\begin{subfigure}[t]{0.48\textwidth}
    \centering
    \includegraphics[width=\linewidth]{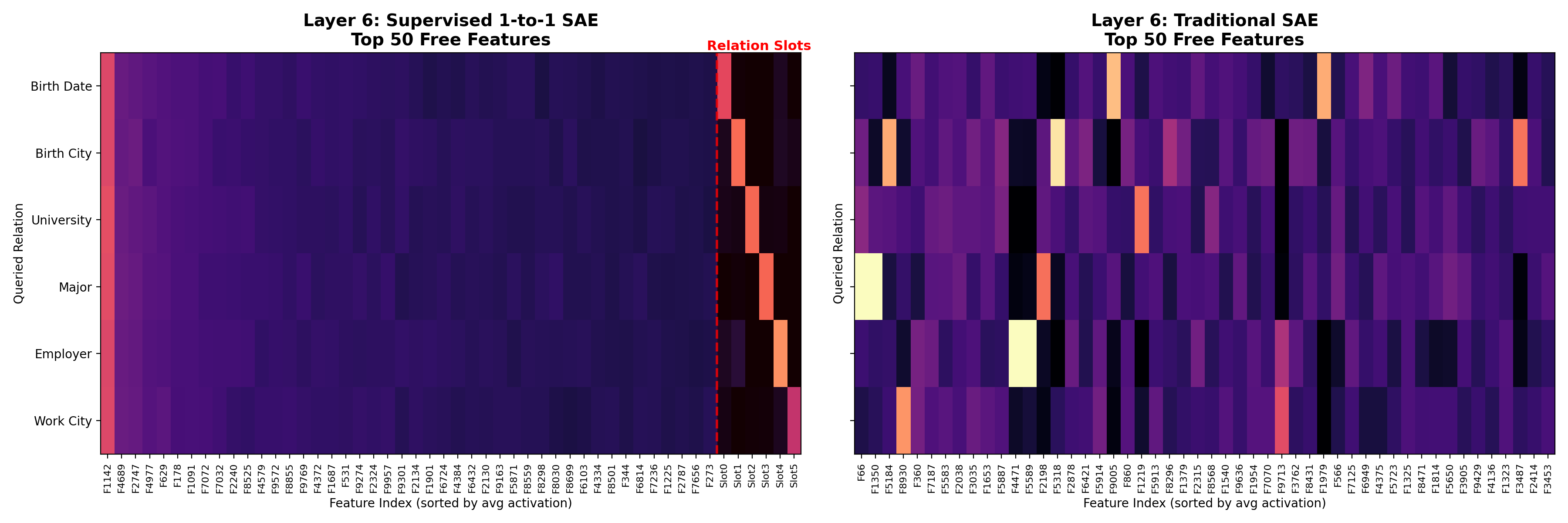}
    \caption{Layer 6}
    \label{fig:sae_layer06}
\end{subfigure}
\hfill
\begin{subfigure}[t]{0.48\textwidth}
    \centering
    \includegraphics[width=\linewidth]{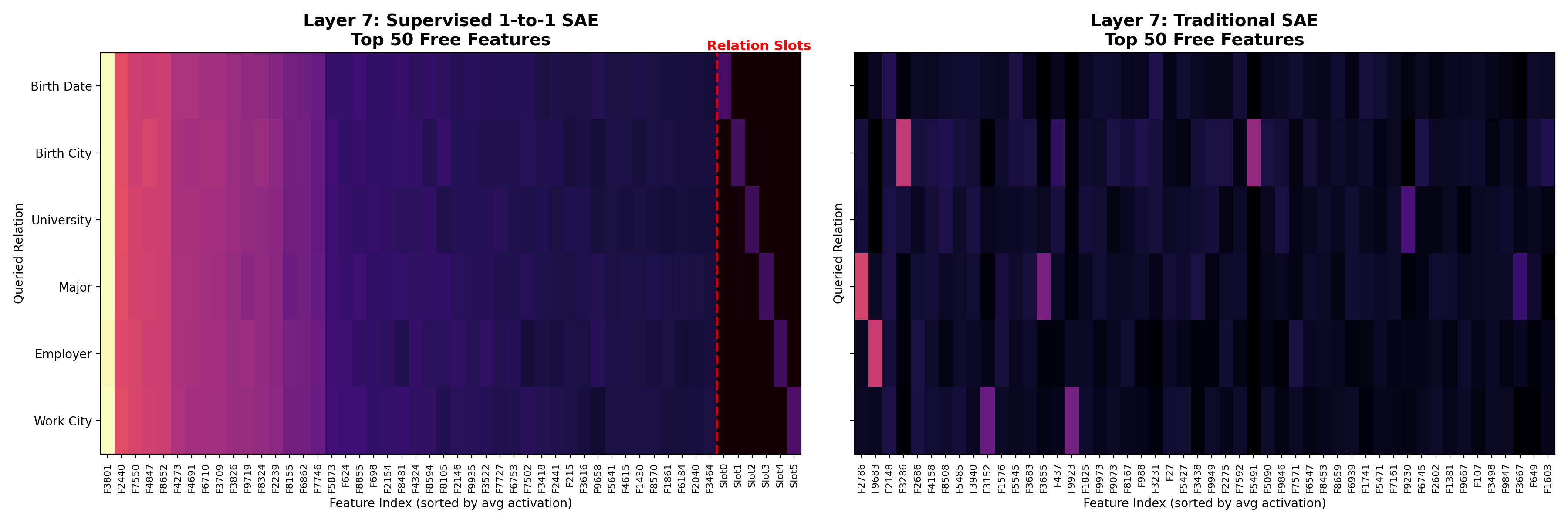}
    \caption{Layer 7}
    \label{fig:sae_layer07}
\end{subfigure}

\vspace{0.15cm}

\begin{subfigure}[t]{0.48\textwidth}
    \centering
    \includegraphics[width=\linewidth]{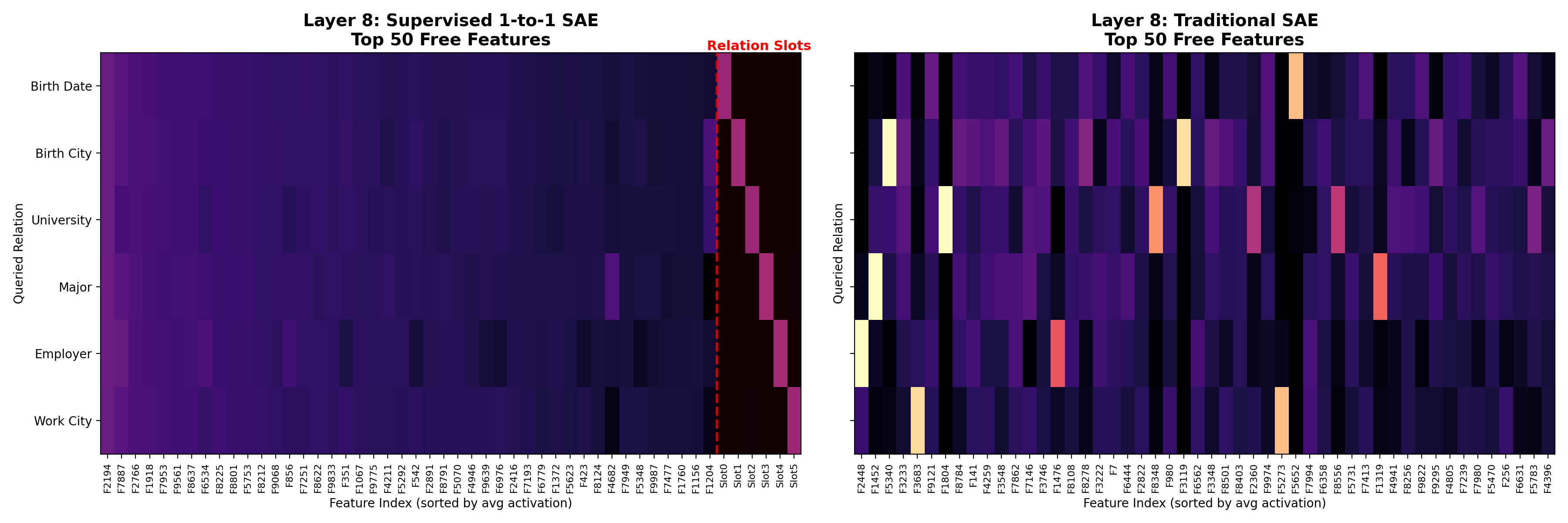}
    \caption{Layer 8}
    \label{fig:sae_layer08}
\end{subfigure}
\hfill
\begin{subfigure}[t]{0.48\textwidth}
    \centering
    \includegraphics[width=\linewidth]{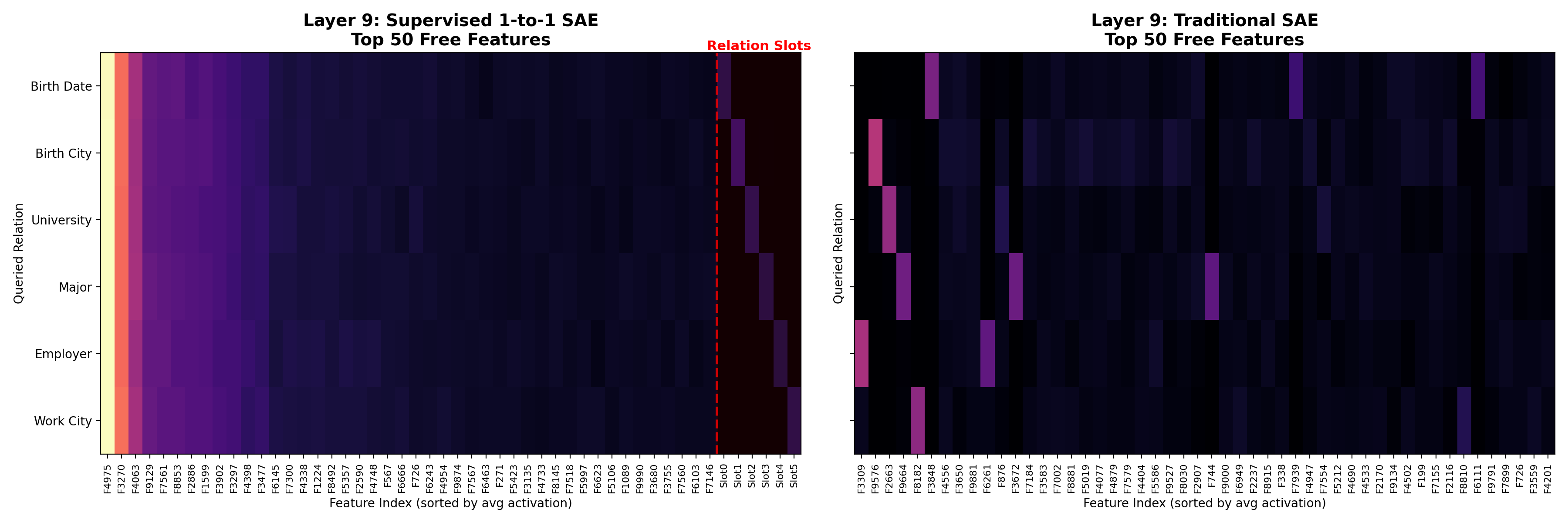}
    \caption{Layer 9}
    \label{fig:sae_layer09}
\end{subfigure}

\vspace{0.15cm}

\begin{subfigure}[t]{0.48\textwidth}
    \centering
    \includegraphics[width=\linewidth]{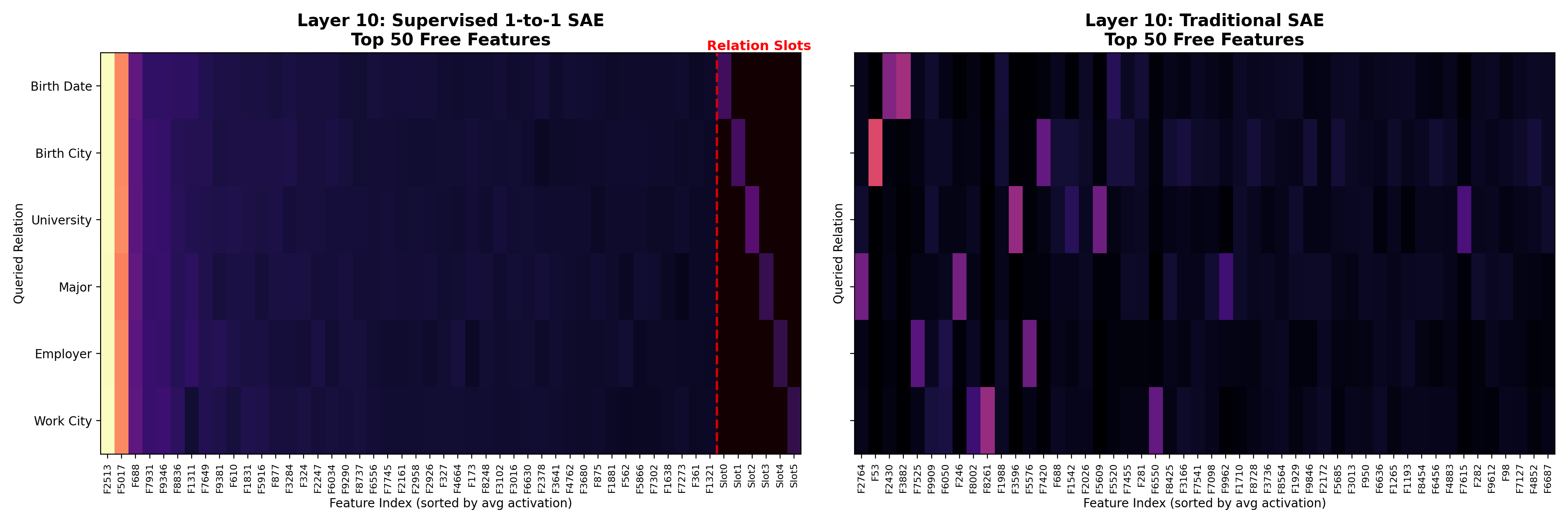}
    \caption{Layer 10}
    \label{fig:sae_layer10}
\end{subfigure}
\hfill
\begin{subfigure}[t]{0.48\textwidth}
    \centering
    \includegraphics[width=\linewidth]{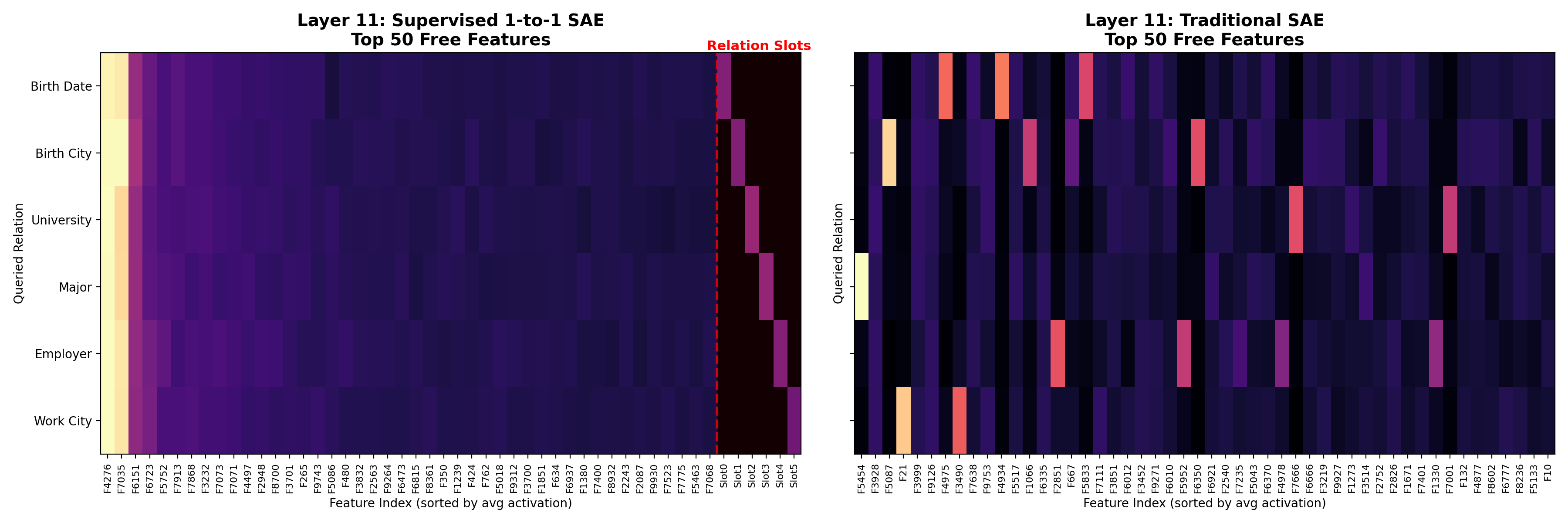}
    \caption{Layer 11}
    \label{fig:sae_layer11}
\end{subfigure}

\caption{Layer-wise top-50 feature activations in GPT-2 with and without SAE post-training. Early layers (0--3) remain diffuse and entangled. Middle layers (4--8) show maximum SAE benefit, with Layer 6 achieving optimal semantic binding and diagonal accuracy. Deep layers (9--11) without SAEs exhibit irregular patterns and over-compression, whereas SAE-trained models maintain structured representations.}
\label{fig:sae_layer_comparison}
\end{figure*}

This section presents a comprehensive comparison of top-50 activated features across all 12 transformer layers (Layer 0--11) of GPT-2 (\cref{fig:sae_layer_comparison}). Each visualization compares two conditions: (1) \textbf{with SAE post-training} (supervised sparse autoencoder applied), and (2) \textbf{without SAE post-training} (baseline activations). This comparison reveals how supervised alignment shapes feature representations and demonstrates the emergence of clean semantic binding in middle layers, while also highlighting artifacts that appear in deeper layers without SAE regularization.

\subsection{Analysis}

These layer-wise visualizations reveal several critical insights about the role of supervised SAE training across network depth:

\paragraph{Early Layers (0--4): Limited Semantic Structure.} In shallow layers, both conditions (with and without SAE) show relatively diffuse activation patterns with weak relation-specific structure. This reflects that early transformer layers primarily process local token-level features and have not yet formed abstract semantic representations suitable for clean concept binding. The SAE provides marginal improvements but cannot overcome the fundamental limitation that these layers lack the representational capacity for high-level semantic concepts.

\paragraph{Middle Layers (5--8): Emergence of Clean Binding.} The most dramatic differences appear in middle layers, particularly Layer 6. With SAE post-training, we observe sharp, diagonal activation patterns indicating successful one-to-one binding between ontological relations and designated slots. Without SAE supervision, these same layers show more scattered, overlapping activations that fail to achieve clean separation between semantic concepts. This demonstrates that while middle layers contain the raw representational power for concept binding, explicit supervision through the SAE's multi-objective loss is essential to crystallize these latent capabilities into interpretable, controllable structure.

\paragraph{Deep Layers (9--11): Artifacts Without SAE.} A particularly interesting phenomenon emerges in deeper layers without SAE post-training. Starting around Layer 6 and becoming more pronounced in Layers 9--11, the baseline (no SAE) condition exhibits strange, irregular activation patterns---potentially including sparse, extreme activations, feature collapse, or unexpected clustering. These artifacts likely reflect the model's aggressive compression of task-relevant information in preparation for final output generation. The supervised SAE mitigates these irregularities by enforcing reconstruction fidelity, sparsity constraints, and orthogonality between relation and free slots, resulting in more stable and interpretable features even in deep layers.

\paragraph{Optimal Layer for Intervention.} These visualizations provide empirical justification for choosing Layer 6 as the primary layer for semantic intervention experiments (as reported in Section~5 of the main paper). Layer 6 achieves: (1) mature semantic representations that support clean binding, (2) strong separation between concepts under SAE training, (3) minimal artifacts compared to deeper layers, and (4) acceptable reconstruction error that preserves model functionality.

\paragraph{The Role of Supervised Training.} Across all layers, SAE post-training consistently produces more structured, interpretable activation patterns. The supervised losses---particularly alignment loss (enforcing relation-slot correspondence) and orthogonality loss (separating supervised and unsupervised features)---act as powerful inductive biases that shape the latent space into a form amenable to human interpretation and causal intervention. Without this supervision, even layers with rich semantic content fail to expose that structure in an accessible format.

\section{Additional Qualitative Swap Examples}
\label{app:qual_swaps}

\begin{figure}[t]
  \centering
  \includegraphics[width=0.5\linewidth]{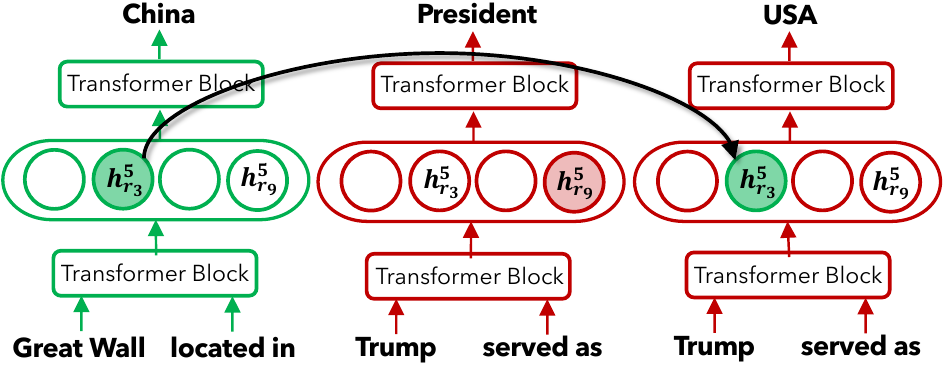}
  \caption{Illustrative schematic of our swap intervention (Orig relation $\rightarrow$ Swap relation).}
  \label{fig:swap}
\end{figure}

We provide qualitative swap examples to complement the quantitative controllability results
(\cref{subsec:swap_controllability}, \cref{subsec:swap_result}). Unless otherwise noted, all examples use Layer~6 and intervention
strength $\alpha{=}2$. In each example, we start from a question whose gold relation is the
\emph{Orig} type and intervene by amplifying the decoded direction of a different concept slot
(\emph{Swap}), aiming to make the model answer with the corresponding target attribute.

\subsection{Swapping Schematic}
\label{app:swap_schematic}

\cref{fig:swap} illustrates the swap intervention: we steer the layer-$\ell$ residual representation along the decoded direction of a target (Swap) concept slot to flip the predicted
answer type from the original relation.

\subsection{Correct Swap Examples}
\label{app:swap_examples_correct}

\begin{table*}[t]
\centering
\setlength{\tabcolsep}{5pt}
\renewcommand{\arraystretch}{1.12}
\footnotesize
\begin{tabularx}{\textwidth}{@{}l X X X@{}}
\toprule
\textbf{Orig $\rightarrow$ Swap} & \textbf{Question} & \textbf{Target} & \textbf{Generated} \\
\midrule
\textsc{company\_city} $\rightarrow$ \textsc{university} &
What is Grace Wendy Rivera's work city? &
Florida International University &
Florida International University \\

\textsc{company\_city} $\rightarrow$ \textsc{major} &
What is Grace Wendy Rivera's work city? &
Electrical Engineering &
Electrical Engineering \\

\textsc{company\_city} $\rightarrow$ \textsc{employer} &
What is Grace Wendy Rivera's work city? &
Blackstone &
Blackstone \\

\textsc{university} $\rightarrow$ \textsc{birth\_date} &
Where did Thomas Heath Stafford go to college? &
2, March, 1981 &
2, March, 1981 \\

\textsc{university} $\rightarrow$ \textsc{major} &
Where did Thomas Heath Stafford go to college? &
Dance &
Dance \\

\textsc{birth\_date} $\rightarrow$ \textsc{work\_city} &
When was Megan Kian Valencia born? &
Framingham, MA &
Framingham, MA \\

\textsc{birth\_city} $\rightarrow$ \textsc{birth\_date} &
Where was Angela Maddox Gates born? &
27, November, 1950 &
27, November, 1950 \\

\textsc{major} $\rightarrow$ \textsc{university} &
What was Jennifer Donovan Pruitt's major? &
University of Wisconsin--Madison &
University of Wisconsin--Madison \\
\bottomrule
\end{tabularx}
\caption{Correct swap examples from the evaluation set (Layer~6; $\alpha{=}2$).}
\label{tab:swap_examples_correct}
\end{table*}

\cref{tab:swap_examples_correct} reports successful swaps. \textbf{Orig$\rightarrow$Swap}
denotes the original relation queried and the target relation we force via intervention.
\textbf{Target} is the gold value $g(p,\text{Swap})$ for the same person $p$, and \textbf{Generated}
is the model output after the swap (counted as correct when it matches the Target).

The swap error analysis, including per-concept category retention statistics and a representative failure case, is presented in \cref{subsec:error_analysis} of the main paper.

\revision{\section{Mutual Information Gap Analysis}
\label{app:mig}

To quantify disentanglement beyond fragmentation and concentration,
we compute the Mutual Information Gap (MIG)~\citep{chen2018isolating}
between supervised slot activations and concept labels.
For each concept $k$, we identify the two slots with highest
mutual information $I(z_j; k)$ and compute}
\begin{equation}
	\mathrm{MIG}_k = \frac{I(z_{j_1}; k) - I(z_{j_2}; k)}{H(k)},
\end{equation}
\revision{where $H(k)$ is the entropy of concept $k$.
Higher MIG indicates that each concept is captured by a single dominant slot.}

\begin{table}[!htbp]
\centering
\begin{tabular}{lccc}
\hline
{\bfseries Setting} & {\bfseries \modelname{}} & {\bfseries Standard SAE} & {\bfseries Ratio} \\
\hline
\revision{1-Hop (6 concepts)}  & \revision{{\bfseries 0.173}} & \revision{0.000} & \revision{$\infty$} \\
\revision{2-Hop (20 relations)} & \revision{{\bfseries 0.035}} & \revision{0.004} & \revision{$9.2\times$} \\
\hline
\end{tabular}
\caption{\revision{Mutual Information Gap (MIG). Standard SAE uses
identical architecture trained with only reconstruction and
sparsity losses. An MIG of 0.000 is expected for the standard
SAE: without concept supervision, no single unsupervised
feature has any incentive to exclusively encode a specific
concept---information is distributed across many features
by design.}}
\label{tab:mig}
\end{table}

\revision{\cref{tab:mig} reports the results: \modelname{} achieves consistent per-concept MIG values ranging
from 0.146 to 0.192 across the 6 concepts (1-hop) and 0.022
to 0.039 across the 20 relations (2-hop). The lower absolute
2-hop MIG reflects the harder 20-way discrimination, but the
$9.2\times$ improvement over the standard SAE demonstrates
substantial disentanglement gains.}

\begin{figure}[t]
\centering
\includegraphics[width=\textwidth]{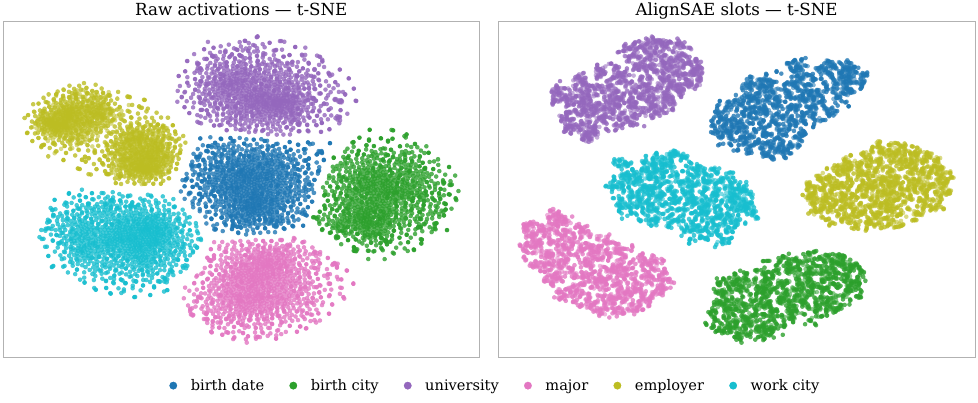}\\[0.8em]
\includegraphics[width=\textwidth]{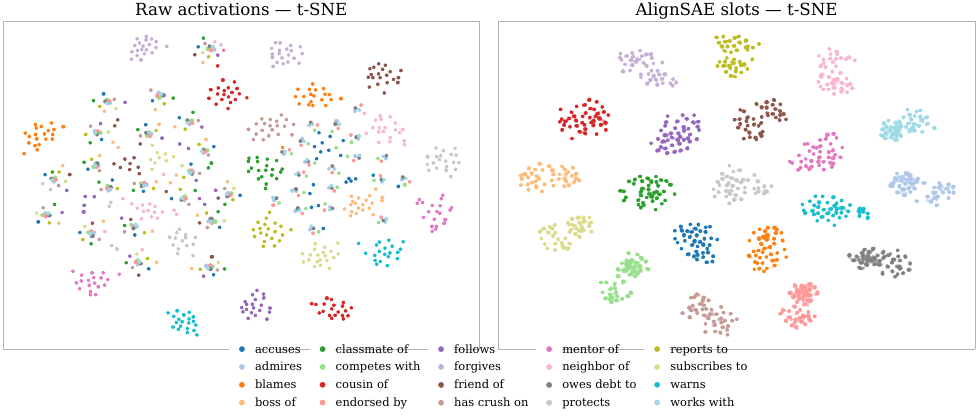}
\caption{\revision{t-SNE projections of raw activations (left)
vs.\ \modelname{} supervised slot activations (right), colored by
concept label. Top: 1-hop (6 concepts). Bottom: 2-hop (20 relations).}}
\label{fig:tsne}
\end{figure}

\begin{figure}[t]
\centering
\includegraphics[width=\textwidth]{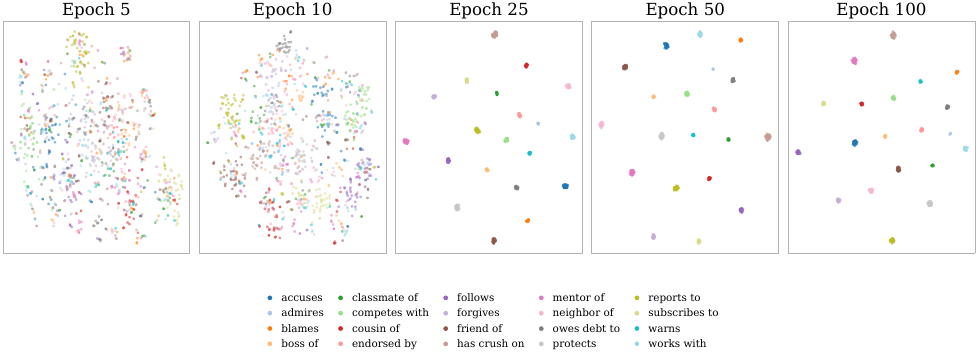}
\caption{\revision{Grokking progression: t-SNE of 2-hop supervised slot
activations at epochs 5, 10, 25, 50, 100.}}
\label{fig:grokking_tsne}
\end{figure}

\revision{\section{Latent Space Visualization (t-SNE)}
\label{app:latent_vis}

Figure~\ref{fig:tsne} visualizes the latent space before and
after alignment via t-SNE projections.
For 1-hop, raw hidden activations (768-dim) show partial but
overlapping clustering by concept; \modelname{} supervised slot
activations (6-dim) produce tight, fully separated clusters
for all 6 concepts.
For 2-hop, raw activations form a largely undifferentiated
cloud; \modelname{} slots create 20 distinct clusters.
Figure~\ref{fig:grokking_tsne} shows the grokking progression:
t-SNE of supervised slot activations at training epochs
5, 10, 25, 50, and 100. At epoch~5, no structure is
visible. By epoch~25, clusters begin forming. By epoch~100,
clean one-to-one separation emerges.}

\begin{figure}[t]
\centering
\includegraphics[width=0.9\textwidth]{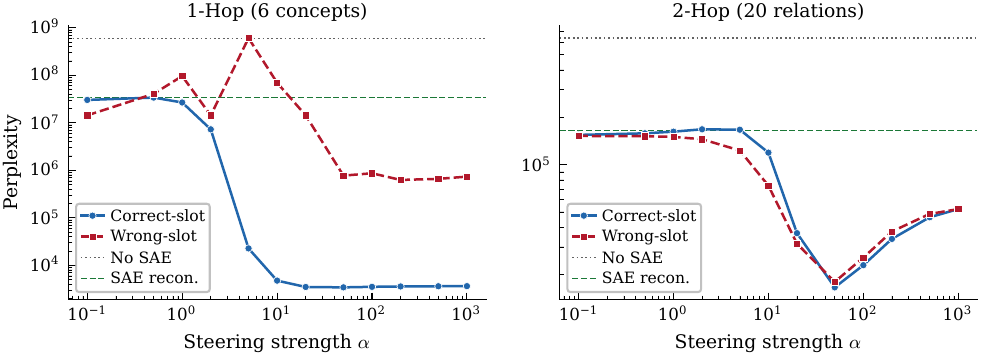}
\caption{\revision{Perplexity vs.\ steering strength $\alpha$.
Correct-slot steering (blue) dramatically reduces perplexity;
wrong-slot steering (red) does not. Dashed lines show
no-SAE and SAE-reconstruct baselines.}}
\label{fig:ppl}
\end{figure}

\revision{\section{Steering Side-Effects Analysis}
\label{app:side_effects}

A practical steering interface must not only induce the target
concept but also preserve other model capabilities. We evaluate
side-effects via perplexity analysis and free-form generation quality.

\paragraph{Perplexity vs.\ Amplification.}
We compute perplexity on held-out QA data after intervening
at different $\alpha$ values, comparing correct-slot steering
(boosting the true concept) vs.\ wrong-slot steering (boosting
an incorrect concept). As shown in \cref{fig:ppl}, correct-slot steering reduces perplexity
by over 5 orders of magnitude from the no-SAE baseline
($\sim$3{,}438 at $\alpha{=}50$ for 1-hop), then plateaus.
Wrong-slot steering maintains high perplexity across all
$\alpha$ values. This asymmetry confirms that \modelname{}
steering provides a precise, targeted signal: it helps when
aligned with the ground truth and hurts when misaligned.
The SAE reconstruction baseline (encode$\to$decode with
no slot modification) already improves perplexity over the
no-SAE condition.}

\revision{\paragraph{Free-Form Generation Quality.}
We evaluate generation from neutral prompts
(``Once upon a time,'', ``The weather today is'', etc.)
with and without steering. Since the model was fine-tuned on
a predefined ontology of structured biographical paragraphs
and QA pairs, it is \emph{designed} to produce ontology-grounded
factual outputs rather than free-form creative text.

\cref{tab:generation_examples} shows representative outputs.
Without steering, the model produces generic biographical text.
When we steer toward a specific concept (birth\_date, major,
or work\_city), the \emph{first generated token} immediately
reflects the target concept---a date, a field of study, or a
city name---while the continuation remains fluent biographical
text. The model's domain-specific behavior is preserved; only
the \emph{leading concept} changes. This first-token effect
demonstrates that \modelname{}'s 1-to-1 concept binding provides
a precise causal handle: the steered slot directly influences
the model's next-token distribution at generation onset.}

\begin{table}[!htbp]
\centering
\small
\setlength{\tabcolsep}{4pt}
\revision{\begin{tabularx}{\textwidth}{lX}
\toprule
\textbf{Condition} & \textbf{First tokens generated} \\
\midrule
No steering &
\textit{Whitman College counts She among its graduates. She works in Purchase, NY. Her field of study was Nursing\ldots} \\
\addlinespace
Steer $\to$ birth\_date ($\alpha{=}10$) &
\textbf{27,July,1991}, \textit{Bryan Caleb Atkins was born. He is professionally active in Fort Worth, TX\ldots} \\
\addlinespace
Steer $\to$ major ($\alpha{=}10$) &
\textbf{French}. \textit{According to records, Maurice Hudson Tran was born on 28,September,1953\ldots} \\
\addlinespace
Steer $\to$ work\_city ($\alpha{=}10$) &
\textbf{Dallas, TX}. \textit{He earned a degree in Nursing. He is a key part of the AT\&T team\ldots} \\
\bottomrule
\end{tabularx}}
\caption{\revision{Free-form generation examples (1-hop, $\alpha{=}10$).
Bold text marks the first tokens generated, which immediately
reflect the steered concept. Continuations remain fluent
biographical text.}}
\label{tab:generation_examples}
\end{table}

\revision{An LLM judge (Claude Haiku 4.5) rated 70 generations across
fluency, coherence, and concept steering success (1--5 scale).
\cref{tab:llm_judge} reports the full results.
Key findings: (i)~steering preserves fluency (baseline
$\approx$ steered: 2.0 vs.\ 2.1 for 1-hop, 4.0 vs.\ 3.9 for
2-hop); (ii)~concept steering success is 3.9/5 for 1-hop;
(iii)~results are consistent across $\alpha{=}10$ and
$\alpha{=}100$. Because the concepts are well isolated via
\modelname{}'s 1-to-1 binding, steering has \emph{fewer undesirable
side-effects}---it changes \emph{what} the model generates
without degrading \emph{how well} it generates.}

\begin{table}[!htbp]
\centering
\small
\setlength{\tabcolsep}{4pt}
\begin{tabular}{llcccc}
\toprule
\textbf{Setting} & \textbf{Condition} & \textbf{Fluency} & \textbf{Coherence} & \textbf{Steering} & \textbf{$n$} \\
\midrule
\multirow{3}{*}{\revision{1-Hop}}
  & \revision{Baseline}         & \revision{2.0} & \revision{1.0} & \revision{---}  & \revision{5} \\
  & \revision{Steered $\alpha{=}10$}  & \revision{2.1} & \revision{1.1} & \revision{3.9} & \revision{15} \\
  & \revision{Steered $\alpha{=}100$} & \revision{2.0} & \revision{1.1} & \revision{3.9} & \revision{15} \\
\midrule
\multirow{3}{*}{\revision{2-Hop}}
  & \revision{Baseline}         & \revision{4.0} & \revision{1.2} & \revision{---}  & \revision{5} \\
  & \revision{Steered $\alpha{=}10$}  & \revision{4.0} & \revision{1.3} & \revision{2.5} & \revision{15} \\
  & \revision{Steered $\alpha{=}100$} & \revision{3.7} & \revision{1.3} & \revision{2.4} & \revision{15} \\
\bottomrule
\end{tabular}
\caption{\revision{LLM-as-Judge evaluation of free-form generation quality (1--5 scale). Steering preserves fluency and coherence while successfully injecting the target concept in 1-hop.}}
\label{tab:llm_judge}
\end{table}

\revision{Since the models are fine-tuned exclusively on synthetic data,
general-purpose benchmarks (e.g., HellaSwag) would not yield
meaningful signal. We adopt a controlled benchmark within the
model's own data distribution: perplexity on held-out QA,
swap success rates, and LLM-as-Judge assessment measure
whether steering disrupts the model's learned capabilities
on its training task.}

\FloatBarrier

\revision{\section{Scalability: FB15K with 91 Relations}
\label{app:fb15k}
To assess scalability beyond small relation sets, we evaluate \modelname{} on a subset of FB15K~\citep{bordes2013translating} with 91 relations.
We retain the same SAE architecture and training protocol, increasing only the supervised slot count to $|\mathcal{R}|{=}91$.
All experiments use Layer~6.

\paragraph{Dataset Construction.}
FB15K contains 91 relation types. For each relation we sample up to 90 unique head entities; relations with fewer than 90 heads retain their original count (see \cref{tab:relation_full} for per-relation statistics).
QA and biography templates are generated with GPT following the same procedure as the 1-hop setting.

\newcommand{\rel}[1]{\makecell[l]{#1}}

\paragraph{Binding and Reconstruction.}

\cref{tab:fb15k_scalability_main} summarizes the binding and reconstruction results.
Train slot top-1 accuracy reaches 98.7\%, and diagonal accuracy is 95.6\% (\cref{fig:fb15k_confusion}), indicating that \modelname{} successfully scales to 91 supervised slots with near-perfect in-distribution binding.
Out-of-distribution slot top-1 accuracy (unseen templates/persons) is 75.7\%, substantially above the $1/91 \approx 1.1\%$ chance level.
Reconstruction MSE remains comparable to the smaller settings.
The two relations with noticeably lower diagonal accuracy---\textit{/base/locations/continents/countries\_within} (4 entities) and \textit{/broadcast/content/artist} (6 entities)---have very few training examples, explaining their weaker performance.

\begin{table}[!htbp]
\centering
\begin{tabular}{l c}
\toprule
\textbf{Metric} & \textbf{Value} \\
\midrule
Train slot top-1 accuracy & 0.987 \\
Test slot top-1 accuracy (OOD) & 0.757 \\
Reconstruction MSE & 0.083 \\
Diagonal accuracy & 0.956 \\
\bottomrule
\end{tabular}
\caption{\modelname{} performance in the 91-relation FB15K setting.}
\label{tab:fb15k_scalability_main}
\end{table}

\begin{figure}[tb]
    \centering
    \includegraphics[width=0.65\linewidth]{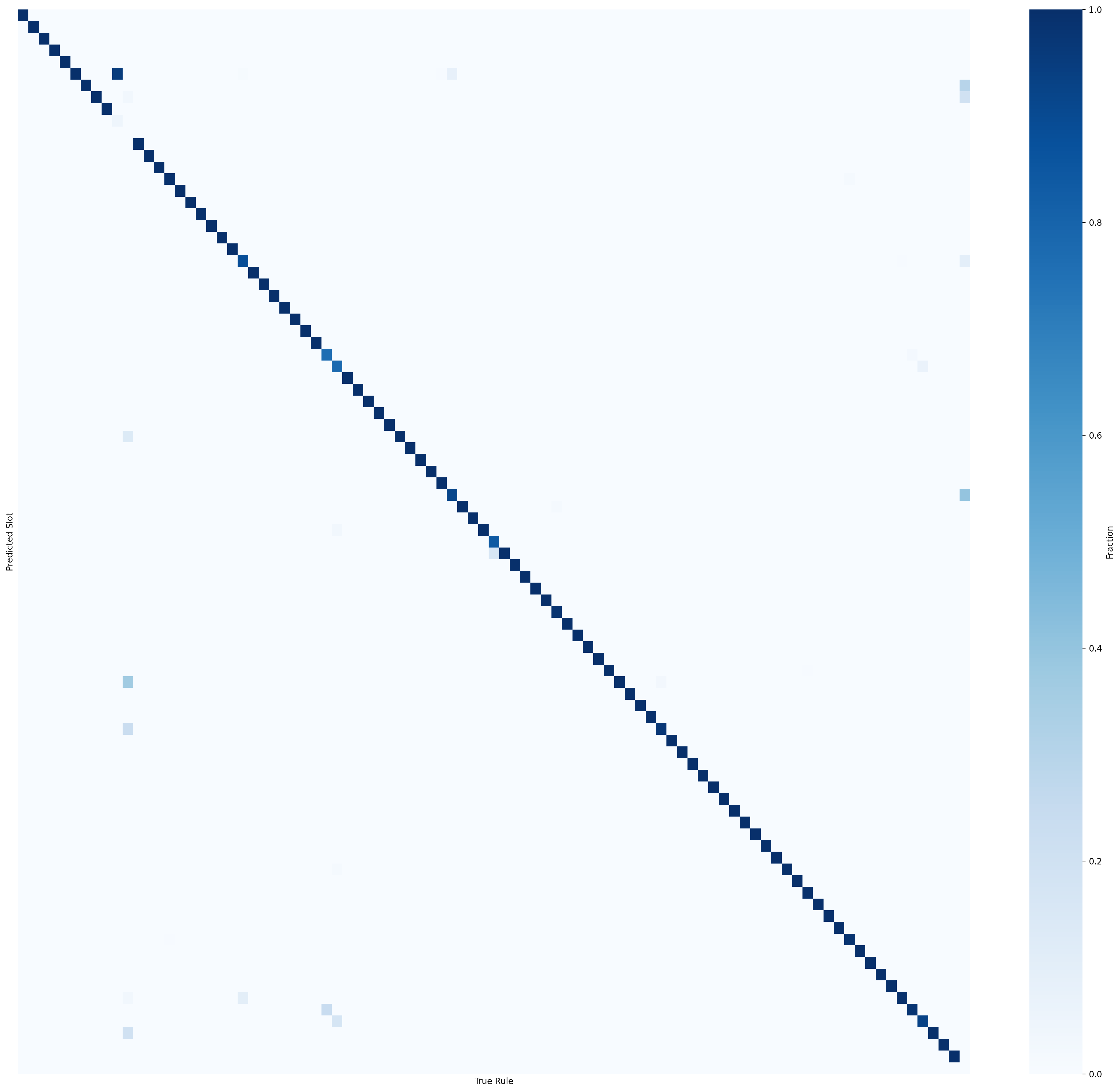}
    \caption{Normalized confusion matrix for 91-relation slot binding at Layer~6. The strong diagonal confirms that \modelname{} maintains near-perfect concept--slot alignment even at this scale.}
    \label{fig:fb15k_confusion}
\end{figure}

\begin{table*}[t]
\centering
\scriptsize
\setlength{\tabcolsep}{1.5pt}

\begin{tabularx}{\textwidth}{r>{\raggedright\arraybackslash}Xr|r>{\raggedright\arraybackslash}Xr}
\toprule
\textbf{Idx} & \textbf{Relation} & \textbf{\#Ent} &
\textbf{Idx} & \textbf{Relation} & \textbf{\#Ent} \\
\midrule

0  & /award/award\_category/category\_of & 90 &
45 & /location/country/second\_level\_divisions & 11 \\

1  & /award/award\_category/disciplines\_or\_subjects & 90 &
46 & /location/hud\_county\_place/county & 90 \\

2  & \rel{/base/aareas/schema/administrative\_area/\\administrative\_area\_type} & 90 &
47 & /location/hud\_county\_place/place & 90 \\

3  & \rel{/base/aareas/schema/administrative\_area/\\administrative\_parent} & 90 &
48 & /location/location/contains & 90 \\

4  & \rel{/base/americancomedy/celebrity\_impressionist/\\celebrities\_impersonated} & 8 &
49 & /location/location/partially\_contains & 90 \\

5  & /base/biblioness/bibs\_location/country & 90 &
50 & /location/location/time\_zones & 90 \\

6  & /base/biblioness/bibs\_location/state & 90 &
51 & /location/us\_county/county\_seat & 90 \\

7  & /base/culturalevent/event/entity\_involved & 65 &
52 & /media\_common/netflix\_genre/titles & 90 \\

8  & /base/eating/practicer\_of\_diet/diet & 90 &
53 & \rel{/medicine/disease/\\notable\_people\_with\_this\_condition} & 22 \\

9  & /base/locations/continents/countries\_within & 4 &
54 & /medicine/disease/risk\_factors & 60 \\

10 & /broadcast/content/artist & 6 &
55 & /medicine/symptom/symptom\_of & 23 \\

11 & /business/business\_operation/industry & 90 &
56 & /music/artist/origin & 90 \\

12 & /dataworld/gardening\_hint/split\_to & 90 &
57 & /music/genre/artists & 90 \\

13 & /education/educational\_institution/campuses & 90 &
58 & /music/genre/parent\_genre & 90 \\

14 & /education/educational\_institution/colors & 90 &
59 & /music/instrument/family & 81 \\

15 & /education/educational\_institution/school\_type & 90 &
60 & /music/instrument/instrumentalists & 86 \\

16 & \rel{/education/educational\_institution\_campus/\\educational\_institution} & 90 &
61 & /music/record\_label/artist & 90 \\

17 & \rel{/education/university/\\fraternities\_and\_sororities} & 90 &
62 & \rel{/olympics/olympic\_games/\\participating\_countries} & 41 \\

18 & /film/director/film & 90 &
63 & /olympics/olympic\_games/sports & 43 \\

19 & /film/film/cinematography & 90 &
64 & /organization/organization/place\_founded & 90 \\

20 & /film/film/costume\_design\_by & 90 &
65 & \rel{/organization/organization\_founder/\\organizations\_founded} & 90 \\

21 & /film/film/country & 90 &
66 & /people/cause\_of\_death/people & 66 \\

22 & /film/film/edited\_by & 90 &
67 & /people/deceased\_person/place\_of\_burial & 90 \\

23 & /film/film/executive\_produced\_by & 90 &
68 & /people/deceased\_person/place\_of\_death & 90 \\

24 & /film/film/featured\_film\_locations & 90 &
69 & /people/ethnicity/geographic\_distribution & 40 \\

25 & /film/film/film\_art\_direction\_by & 86 &
70 & /people/ethnicity/languages\_spoken & 70 \\

26 & /film/film/film\_festivals & 90 &
71 & /people/ethnicity/people & 75 \\

27 & /film/film/film\_format & 90 &
72 & /people/person/gender & 90 \\

28 & /film/film/film\_production\_design\_by & 90 &
73 & /people/person/languages & 90 \\

29 & /film/film/genre & 90 &
74 & /people/person/nationality & 90 \\

30 & /film/film/language & 90 &
75 & /people/person/place\_of\_birth & 90 \\

31 & /film/film/music & 90 &
76 & /people/person/profession & 90 \\

32 & /film/film/prequel & 90 &
77 & /people/person/religion & 90 \\

33 & /film/film/produced\_by & 90 &
78 & /people/profession/specialization\_of & 90 \\

34 & /film/film/production\_companies & 90 &
79 & /sports/sports\_team/colors & 90 \\

35 & /film/film/story\_by & 90 &
80 & /sports/sports\_team/sport & 90 \\

36 & /film/film/written\_by & 90 &
81 & /sports/sports\_team\_location/teams & 90 \\

37 & \rel{/film/film\_set\_designer/\\film\_sets\_designed} & 24 &
82 & \rel{/time/event/instance\_of\_recurring\_event} & 90 \\

38 & /film/film\_subject/films & 90 &
83 & /time/event/locations & 90 \\

39 & /influence/influence\_node/influenced\_by & 90 &
84 & /tv/tv\_program/country\_of\_origin & 90 \\

40 & /language/human\_language/countries\_spoken\_in & 56 &
85 & /tv/tv\_program/genre & 90 \\

41 & /location/administrative\_division/country & 90 &
86 & /tv/tv\_program/languages & 90 \\

42 & /location/country/capital & 90 &
87 & /tv/tv\_program/program\_creator & 90 \\

43 & /location/country/form\_of\_government & 90 &
88 & \rel{/user/alexander/philosophy/philosopher/\\interests} & 39 \\

44 & /location/country/official\_language & 90 &
89 & \rel{/user/jg/default\_domain/olympic\_games/\\sports} & 40 \\

\bottomrule
\end{tabularx}

\caption{All 91 FB15K relations with the number of unique head entities per relation. Relations with fewer than 90 entities retain their original count.}
\label{tab:relation_full}
\end{table*}
}

\revision{\section{Standard SAE Baseline and 2-Hop Controllability Gap}
\label{app:twohop_gap}

\paragraph{Standard SAE Baseline.}
Using the same swap-strength sweep shown in Figure~\ref{fig:2hop_swap},
we compare \modelname{} against a standard (unsupervised) SAE with identical
architecture trained only with reconstruction and sparsity losses.
The standard SAE remains near floor-level control across $\alpha$
($\sim$1.5--2.7\%), while \modelname{} reaches 7.86\% at its best
setting ($\alpha\approx50$), indicating that supervised alignment is
necessary for meaningful steerability in 2-hop settings.

\paragraph{The 2-Hop Controllability Gap.}
While \modelname{} achieves near-perfect swap success in 1-hop
(94.8\%), the 2-hop rate of 7.86\% remains low in absolute
terms, despite being $2.9\times$ higher than the standard SAE.
We attribute this gap to the compositional nature of 2-hop
reasoning: the intervention modifies the hidden state at a
single position in a single layer, but the model's computation
of $e_3$ depends on attention patterns distributed across
\emph{all} previous token positions. A single-position
intervention cannot fully redirect this distributed computation.
Crucially, diagonal accuracy remains near-perfect (100\%,
\cref{fig:2hop_binding_grokking_full}), revealing that
\emph{binding $\neq$ full causal control} in multi-hop settings.}

\revision{\section{Applications and Future Directions}
\label{app:future}

\paragraph{Applications.}
Concept-aligned slots enable several downstream applications:}
\revision{(i)~\emph{targeted model editing}---modifying a specific
fact by intervening on the corresponding slot, as
demonstrated by our swap experiments (94.8\% success);
(ii)~\emph{interpretable monitoring}---observing which
concept slots activate during inference to audit what
knowledge the model accesses;
(iii)~\emph{bias detection}---examining whether certain
concept slots systematically co-activate with others;
and (iv)~\emph{knowledge diagnostics}---using per-concept
diagonal accuracy to identify gaps in a model's learned
knowledge. Natural extensions to real-world data include
Wikidata relation triples, biomedical knowledge graphs
(Gene Ontology, UMLS), and financial ontologies (FIBO).

\paragraph{Future Directions.}
\modelname{} operates over a predefined ontology of discrete
concepts, matching the structure of most real-world knowledge
domains---in biomedicine, finance, and law, concepts are
deterministic entries in structured ontologies, not fuzzy
categories. Scaling \modelname{} to larger ontologies
is a natural next step;
the architecture scales linearly ($K$ additional
encoder/decoder columns per concept).
As a first validation, we evaluate on FB15K with 91 relations
(\cref{app:fb15k}), achieving 98.7\% train slot top-1 accuracy
and 95.6\% diagonal accuracy---demonstrating that concept
alignment remains effective well beyond the original 6- and
20-concept settings.
Extending further to thousands of concepts and
to real-world corpora (Wikidata, Gene Ontology, UMLS)
is a concrete direction.
The 2-hop controllability gap (7.86\% swap success vs.\
94.8\% in 1-hop) motivates multi-position and multi-layer
intervention strategies.
Additionally, we envision extending to larger
model backbones, hierarchical concept ontologies, and
cross-modal architectures.}

\section{Two-hop Reasoning Data Format}
\label{app:twohop_format}

\paragraph{Dataset and setup.}
We further evaluate \modelname{} on a \emph{two-hop} reasoning task adapted from prior work~\citep{du-etal-2025-reason}. Each instance specifies a starting entity $e_1$ and an ordered pair of relations $(r_1,r_2)$, and requires predicting a two-step chain $(e_2,e_3)$ that forms the compositional path
$e_1 \xrightarrow{r_1} e_2 \xrightarrow{r_2} e_3$,
\textit{i.e.}, $e_2 = r_1(e_1)$ and $e_3 = r_2(e_2)$.
We instantiate the two-hop ontology with 20 relation types:
\texttt{accuses}, \texttt{admires}, \texttt{blames}, \texttt{boss\_of}, \texttt{classmate\_of},
\texttt{competes\_with}, \texttt{cousin\_of}, \texttt{endorsed\_by}, \texttt{follows},
\texttt{forgives}, \texttt{friend\_of}, \texttt{has\_crush\_on}, \texttt{mentor\_of},
\texttt{neighbor\_of}, \texttt{owes\_debt\_to}, \texttt{protects}, \texttt{reports\_to},
\texttt{subscribes\_to}, \texttt{warns}, \texttt{works\_with}.

Queries follow compositional templates such as “Who is the $r_2$ of the $r_1$ of $e_1$?”, and are accompanied by short profile sentences that verbalize the hop-1 and hop-2 facts needed to answer the question. Concretely, each instance includes (i) hop~1 evidence describing the triple $(e_1, r_1, e_2)$ and (ii) hop~2 evidence describing $(e_2, r_2, e_3)$, each realized as multiple natural-language paraphrases (see \cref{fig:twohop_example_box}).

We impose \emph{step-wise} supervision (illustrated in \cref{fig:2hop_overview}): the model is trained to output the intermediate entity followed by the final entity (\textit{i.e.}, ``$e_2\;e_3$''), so that slot alignment can be evaluated at each hop. Concretely, when generating $e_2$ the aligned concept slot should indicate $r_1$, and when generating $e_3$ it should indicate $r_2$.
We generate 8{,}000 question--answer pairs in total and use a 4{,}000/4{,}000 train/validation split (no label noise).
Based on the layer sweep in \cref{subsec:swap_controllability}, we use layer~6 for all two-hop experiments.

\paragraph{Example.}
\cref{fig:twohop_example_box} shows one dataset instance. The question asks for the $R_2$-relation of the $R_1$-relation of the starting entity (\textit{Avery}). The supporting profile sentences for hop~1 state that \textit{Dominic} \textit{reports\_to} \textit{Avery}, and the hop~2 profile sentences state that \textit{Gerald} is a \textit{friend\_of} \textit{Dominic}. The step-wise target output is therefore \textit{``Dominic Gerald''}.

\begin{figure*}[t]
\vspace{-0.4em}
\centering
\setlength{\fboxsep}{10pt}
\noindent\fbox{%
\begin{minipage}{\dimexpr\linewidth-2\fboxsep-2\fboxrule\relax}
\small
\textbf{Two-hop example (step-wise supervision).}\vspace{0.5em}

\begin{tabularx}{\linewidth}{@{}lX@{}}
\toprule
\textbf{Query} & Who is the \texttt{friend} of the \texttt{report} of \texttt{Avery}? \\
\textbf{Canonical} & $r_1=\texttt{reports\_to}$,\;\; $r_2=\texttt{friend\_of}$ \\
\textbf{Target} & \texttt{Dominic Gerald}\;\; (\textit{i.e.}, $e_2=\texttt{Dominic}$,\; $e_3=\texttt{Gerald}$) \\
\bottomrule
\end{tabularx}

\vspace{0.65em}

\begin{tabularx}{\linewidth}{@{}X X@{}}
\textbf{Hop 1 evidence ($e_1 \xrightarrow{r_1} e_2$)} &
\textbf{Hop 2 evidence ($e_2 \xrightarrow{r_2} e_3$)}\\
\texttt{Dominic} \texttt{reports\_to} \texttt{Avery} &
\texttt{Gerald} \texttt{friend\_of} \texttt{Dominic} \\
\begin{itemize}[leftmargin=1.1em,itemsep=0.12em,topsep=0.12em]
  \item {``Dominic needs to send regular updates to Avery.''}
  \item {``Dominic works under the supervision of Avery.''}
  \item {``Avery is Dominic's manager at work.''}
  \item {``Dominic answers directly to Avery.''}
  \item {``Avery oversees Dominic's work tasks.''}
  \item {``Dominic's reports go straight to Avery.''}
\end{itemize}
&
\begin{itemize}[leftmargin=1.1em,itemsep=0.12em,topsep=0.12em]
  \item {``Gerald and Dominic are best pals.''}
  \item {``Everyone knows that Gerald is a close friend of Dominic.''}
  \item {``Gerald often hangs out with Dominic.''}
  \item {``Dominic and Gerald are good friends.''}
  \item {``Gerald is friendly with Dominic.''}
  \item {``Gerald considers Dominic a trusted friend.''}
\end{itemize}
\end{tabularx}

\vspace{0.5em}
\textbf{\textit{Aligned-slot supervision:}} when predicting $e_2$, the slot indicates $r_1$; when predicting $e_3$, it indicates $r_2$.
\end{minipage}%
}
\vspace{-0.6em}
\caption{\textbf{Two-hop dataset example.} Input query, step-wise targets, and hop-specific profile evidence.}
\label{fig:twohop_example_box}
\vspace{-0.9em}
\end{figure*}

\paragraph{Training hyperparameters.}
We train the base GPT-2 (124M; 12 layers, 768-d) with AdamW and a warmup+cosine learning-rate schedule.
We did \textbf{not} conduct an extensive hyperparameter search; instead, we adopt a {widely used}
configuration for GPT-style fine-tuning (AdamW with weight decay, linear warmup, cosine decay, and moderate
batch sizes) and keep it fixed across runs, only sanity-checking for stable convergence (no divergence).
Concretely, we use an effective batch size of 96, maximum sequence length 512, and train for 100{,}000
update steps with learning rate warmed up for 2{,}000 steps to a peak of $5\times 10^{-4}$ and then
cosine-decayed to a minimum of $1\times 10^{-5}$. We set weight decay to 0.1 and Adam $\epsilon$ to
$1\times 10^{-8}$. The training performance for the two-hop task is illustrated in \cref{fig:twohop_training_curves_grokking}.

\section{Two-Hop Training Dynamics}
\label{app:grokking}

\begin{figure}[htbp]
    \includegraphics[width=0.72\linewidth]{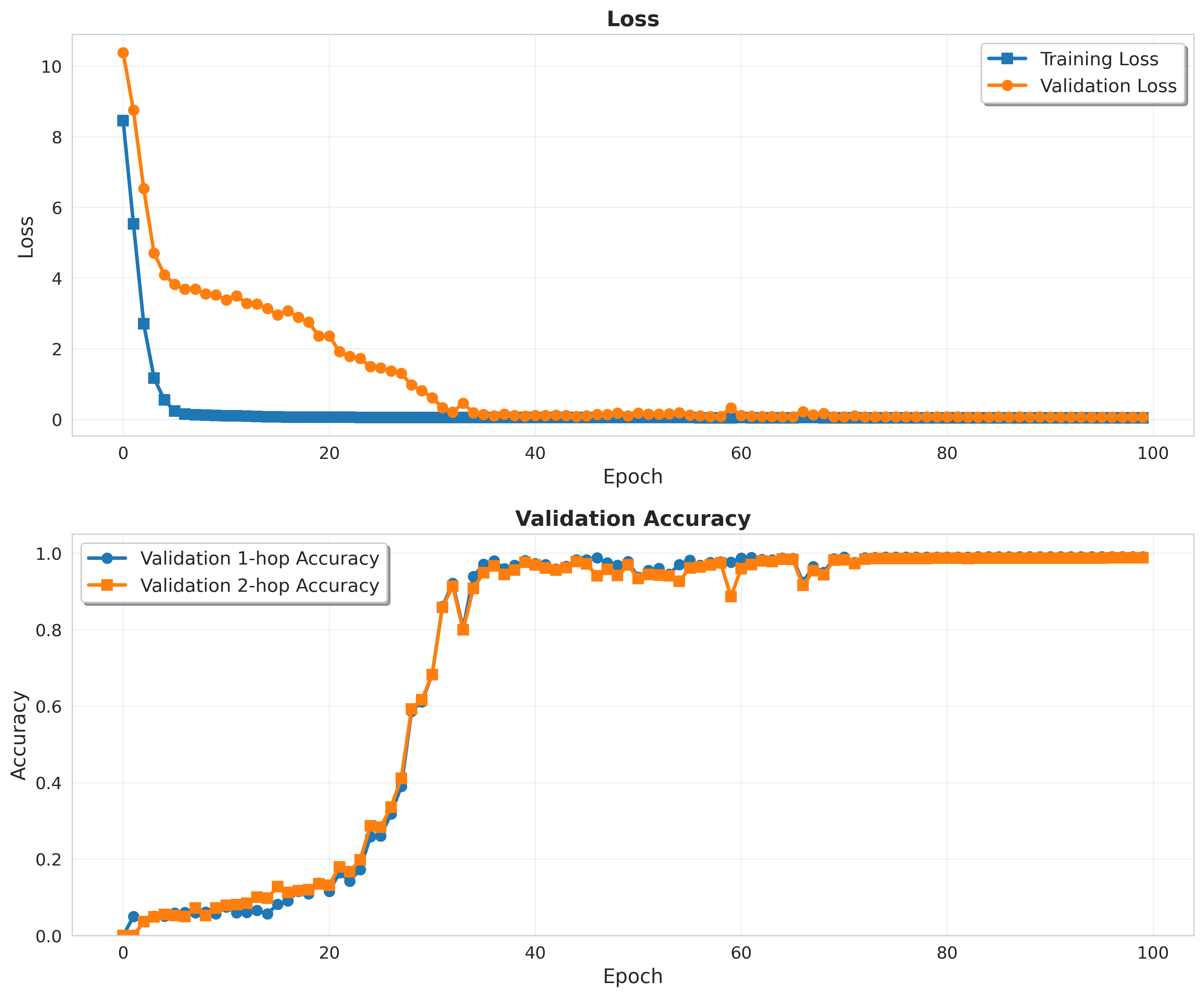}
    \caption{Training dynamics on the two-hop task (base model). Top: training vs.\ validation loss.
    Bottom: validation accuracy for the {1st hop} (Token~1, predicting the intermediate entity $e_2$) and the {2nd hop} (Token~2, predicting the final entity $e_3$) under step-wise supervision.}
    \label{fig:twohop_training_curves_grokking}
\end{figure}

\cref{fig:twohop_training_curves_grokking} shows the full training dynamics of the base two-hop model. Loss decreases smoothly, while validation accuracy remains low for many epochs before a sharp transition, exhibiting the characteristic grokking pattern. The main grokking analysis---including the binding-versus-accuracy dissociation and pre/post-grokking confusion matrices---is presented in \cref{sec:2hop}.

\end{document}